\documentclass{article}

\usepackage{microtype}
\usepackage{graphicx}
\usepackage{subcaption}
\usepackage{booktabs} % for professional tables

\usepackage{hyperref}

 \usepackage[accepted]{icml2026}

\usepackage{amsmath}
\usepackage{amssymb}
\usepackage{mathtools}
\usepackage{amsthm}

\usepackage[capitalize,noabbrev]{cleveref}

\usepackage[textsize=tiny]{todonotes}

\icmltitlerunning{Flow Matching Calibration for Simulation-Based Inference under Model Misspecification}
\usepackage{enumitem}
\usepackage{subcaption}
\usepackage[normalem]{ulem}
\usepackage{booktabs}
\usepackage{multirow}
\usepackage{caption}
\usepackage{adjustbox}

\usepackage{pgfplots}
\pgfplotsset{compat=1.18}
\usepgfplotslibrary{groupplots}

\usepackage{graphicx}
\definecolor{best}{RGB}{0,120,0}
\definecolor{cNPE}{RGB}{150,150,150}   % NPE / FMPE (uncalibrated)
\definecolor{cA}{RGB}{31,119,180}      % FMCPE
\definecolor{cC}{RGB}{44,160,44}       % DPE
\definecolor{cD}{RGB}{214,39,40}       % MF-NPE
\definecolor{cE}{RGB}{148,103,189}     % RoPE

\pgfplotsset{
  methodplot/.style={
    width=0.48\textwidth, height=5.5cm,
    xlabel={$n_{\text{cal}}$},
    xmode=log, log basis x=10,
    xtick={10,50,200,1000},
    xticklabels={10,50,200,1000},
    legend style={font=\footnotesize, at={(0.02,0.98)}, anchor=north west},
    grid=major, grid style={gray!30},
    every axis plot/.append style={thick, mark size=2.5pt},
  }
}

\usepackage{amsmath,amsfonts,bm}

\def\eqref#1{equation~\ref{#1}}
\def\1{\bm{1}}

\def\rvtheta{{\bm{\theta}}}

\def\rvx{{\bm{x}}}
\def\rvy{{\bm{y}}}

\def\rvTheta{{\bm{\Theta}}}

\def\rvX{{\bm{X}}}

\DeclareMathAlphabet{\mathsfit}{\encodingdefault}{\sfdefault}{m}{sl}
\SetMathAlphabet{\mathsfit}{bold}{\encodingdefault}{\sfdefault}{bx}{n}

\newcommand{\R}{\mathbb{R}}

\newcommand{\dd}{\mathrm{d}}

\newcommand{\ncal}{N_{\text{cal}}}

\newcommand{\priordim}{p}

\newcommand{\thetaspace}{\mathbb{R}^{\priordim}}

\newcommand{\pth}{p_{\bm{\Theta}}}
\newcommand{\pxth}{p_{\bm{X}|\bm{\Theta}}}

\newcommand{\pthx}{p_{\bm{\Theta}|\bm{X}}}

\newcommand{\pthy}{p_{\bm{\Theta}|\bm{Y}}}
\newcommand{\pthyv}{p_{\bm{\Theta}|\bm{Y}}(\bm{\theta} | \bm{y})}
\newcommand{\pxy}{p_{\scriptstyle \bm{X}|\bm{Y}}}

\newcommand{\proposal}{\pi_{\bm{\Theta}|\bm{Y}}}
\newcommand{\proposalsimp}{\pi_{\bm{\Theta}|\bm{Y}}}

\newcommand{\ux}{u_{\bm{X}}}
\newcommand{\uth}{u_{\bm{\Theta}}}
\newcommand{\hux}{\hat{u}_{\bm{X}}}
\newcommand{\huth}{\hat{u}_{\bm{\Theta}}}

\RequirePackage{amsthm,amsmath,amsfonts,amssymb}
\RequirePackage{xspace}
\def\XS{\xspace}
\DeclareMathAlphabet{\mathb}{OML}{cmm}{b}{it}
\def\sbmm#1{\ensuremath{\boldsymbol{#1}}}          % Style gras italique (necessite amsmath)
\def\thetab      {{\sbmm{\theta}}\XS}

\newcommand{\xv}{\ensuremath{\bm{x}}}

\newcommand{\yv}{\ensuremath{\bm{y}}}

\newcommand{\ROPE}{{RoPE}}
\newcommand{\ourmethod}{{FMCPE}}
\newcommand{\MFNPE}{{MFNPE}}
\newcommand{\TaskA}{{{Gaussian}}}
\newcommand{\TaskB}{{{Pendulum}}}
\newcommand{\TaskC}{{{Wind~Tunnel}}}
\newcommand{\TaskD}{{{Light~Tunnel}}}
\begin{document}

\twocolumn[
  \icmltitle{Flow Matching Calibration for Simulation-Based Inference under Model Misspecification}
 %\icmltitle{Flow Matching Calibration for Simulation-Based Inference {of Bayesian Models under} Misspecification}

  % It is OKAY to include author information, even for blind submissions: the
  % style file will automatically remove it for you unless you've provided
  % the [accepted] option to the icml2026 package.

  % List of affiliations: The first argument should be a (short) identifier you
  % will use later to specify author affiliations Academic affiliations
  % should list Department, University, City, Region, Country Industry
  % affiliations should list Company, City, Region, Country

  % You can specify symbols, otherwise they are numbered in order. Ideally, you
  % should not use this facility. Affiliations will be numbered in order of
  % appearance and this is the preferred way.
  \icmlsetsymbol{equal}{*}

  \begin{icmlauthorlist}
    \icmlauthor{Pierre-Louis Ruhlmann}{1}
    \icmlauthor{Michael Arbel}{1}
    \icmlauthor{Florence Forbes}{1}
     \icmlauthor{Pedro L. C. Rodrigues}{1}
   % \icmlauthor{Firstname5 Lastname5}{yyy}
  %  \icmlauthor{Firstname6 Lastname6}{sch,yyy,comp}
  %  \icmlauthor{Firstname7 Lastname7}{comp}
    %\icmlauthor{}{sch}
    %\icmlauthor{Firstname8 Lastname8}{sch}
   % \icmlauthor{Firstname8 Lastname8}{yyy,comp}
    %\icmlauthor{}{sch}
    %\icmlauthor{}{sch}
  \end{icmlauthorlist}

  \icmlaffiliation{1}{Univ. Grenoble Alpes, Inria, CNRS, Grenoble INP, LJK, 38000 Grenoble, France}
 % \icmlaffiliation{2}{Company Name, Location, Country}
  %\icmlaffiliation{sch}{School of ZZZ, Institute of WWW, Location, Country}

\icmlcorrespondingauthor{Firstname Lastname}{firsname.lastname@inria.fr}
  %\icmlcorrespondingauthor{Pierre-Louis Ruhlmann}{pierre-louis.ruhlmann@inria.fr}
 % \icmlcorrespondingauthor{Pedro Coelho Rodrigues}{pedro.rodrigues@inria.fr}

  % You may provide any keywords that you find helpful for describing your
  % paper; these are used to populate the "keywords" metadata in the PDF but
  % will not be shown in the document
  \icmlkeywords{Simulation-Based Infererence, Flow Matching, Posterior Estimation, Multi-Fidelity, Model Misspecification}

  \vskip 0.3in
]
% This command actually creates the footnote in the first column listing the
% affiliations and the copyright notice. The command takes one argument, which
% is text to display at the start of the footnote. The \icmlEqualContribution
% command is standard text for equal contribution. Remove it (just {}) if you
% do not need this facility.

% Use ONE of the following lines. DO NOT remove the command.
% If you have no special notice, KEEP empty braces:
\printAffiliationsAndNotice{}  % no special notice (required even if empty)
% Or, if applicable, use the standard equal contribution text:
% \printAffiliationsAndNotice{\icmlEqualContribution}
\begin{abstract}
Simulation-based inference (SBI) is transforming experimental sciences by enabling parameter estimation in complex non-linear models from simulated data. A persistent challenge, however, is model misspecification.
In a Bayesian setting, targeting posterior distributions, errors may arise from the simulator, the noise or prior modelling. These model components are only approximations of reality, and severe mismatches can yield biased or overconfident posteriors. We address this issue by introducing Flow Matching Corrected Posterior Estimation (\ourmethod), a framework that leverages the flow matching paradigm to refine simulation-trained posterior estimators using a small set of calibration samples. Our approach proceeds in two stages: first, a posterior approximator is trained on abundant simulated data; second, flow matching transports its predictions toward the true posterior supported by {calibration} observations.
{We rely on the later to guide the correction,} without requiring  explicit knowledge of the misspecification {form or of which model components are affected}. This design enables \ourmethod~to combine the scalability of SBI with robustness to distributional shift. Across synthetic benchmarks and real-world datasets, we show that our proposal consistently mitigates the effects of misspecification, delivering improved inference accuracy and uncertainty {quantification} compared to standard SBI baselines, while remaining computationally efficient.
\end{abstract}

%\begin{abstract}
%Simulation-based inference (SBI) is transforming experimental sciences by enabling parameter estimation in complex non-linear models from simulated data. A persistent challenge, however, is model misspecification: simulators are only approximations of reality, and mismatches between simulated and real data can yield biased or overconfident posteriors. We address this issue by introducing Flow Matching Corrected Posterior Estimation (\ourmethod), a framework that leverages the flow matching paradigm to refine simulation-trained posterior estimators using a small set of calibration samples. Our approach proceeds in two stages: first, a posterior approximator is trained on abundant simulated data; second, flow matching transports its predictions toward the true posterior supported by real observations, without requiring explicit knowledge of the misspecification. This design enables \ourmethod~to combine the scalability of SBI with robustness to distributional shift. Across synthetic benchmarks and real-world datasets, we show that our proposal consistently mitigates the effects of misspecification, delivering improved inference accuracy and uncertainty calibration compared to standard SBI baselines, while remaining computationally efficient.
%\end{abstract}

\section{Introduction}

Many fields of science and engineering describe complex phenomena using {stochastic models, 
capturing inherent sources of randomness, such as measurement noise and probabilistic dynamics. In a Bayesian setting, these models can be decomposed into two components, a likelihood over data $\bm{x} \in \mathbb{R}^d$ and a prior distribution over some parameters of interest $\bm{\theta} \in \mathbb{R}^p$.
%and a noise model that can take care of other sources of randomness or nuisance parameters gathered in $\bm{\epsilon} \in \mathbb{R}^q$. The distinction between $\bm{\theta}$ and $\bm{\epsilon}$ being not always necessary depending on applications \cite{schmitt_detecting_2024}. 
Mechanistic simulators are then often used  to produce synthetic data 
$\bm{x}$ given parameters $\bm{\theta}$, but they rarely yield tractable likelihoods, making classical Bayesian} inference methods such as Markov chain Monte Carlo (MCMC) \citep{mcmc_robert_casella} or variational inference \citep{pmlr-v37-rezende15} inapplicable. Simulation-Based Inference (SBI)~\citep{Deistler2025, cranmer_frontier_2020} addresses this limitation by performing Bayesian parameter inference directly from simulated datasets, bypassing the need for an explicit likelihood $p_{\bm{X}|\bm{\Theta}}$.

\begin{figure}[hbtp]
    \centering
    \includegraphics[width=\columnwidth]{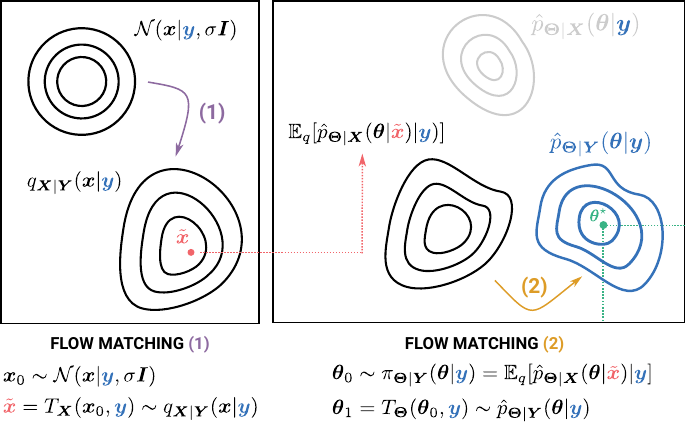}
    \caption{
    \footnotesize
    % Our method \ourmethod\ relies on two interacting flow matching procedures (1) and (2) acting to mitigate the effects of misspecification and avoid innacurate parameter inference with the simulation-based posterior estimate $\hat{p}_{\bm{\Theta}|\bm{X}}$ (represented as \textbf{\color{gray}grey} level sets). In (1) the scarce ground truth calibration data are used to learn a transport $T_{\bm{X}}$ taking real observations $\rvy$ to their counterparts $\tilde{\rvx}$ in the same support as the simulated data. We use a $q_{\bm{X}|\bm{Y}}$-weighted version of $p_{\bm{\Theta}|\bm{X}}$ conditioned on $\tilde{\bm{x}}$ as source distribution for the continuous normalizing flow in (2) and learn a transport $T_{\bm{\Theta}}$ that yields samples from our final approximator $\hat{p}_{\bm{\Theta}|\bm{Y}}$. Note that both (1) and (2) are necessary to balance imperfections on different parts of the pipeline.
    Overview of \ourmethod. The method combines two complementary flow matching steps to correct simulation-based posterior distributions under model misspecification (represented by $\hat{p}_{\bm{\Theta}|\bm{X}}(\bm{\theta}|\bm{y})$ in \textbf{\color{gray}grey} level sets). (1) Scarce calibration data $(\bm{\theta}, \bm{y})$ are used to learn a transport map $T_{\bm{X}}$ that couples real observations $\bm{y}$ with surrogate counterparts $\tilde{\bm{x}}$ lying in the simulator's domain. (2) We then learn $T_{\bm{\Theta}}$ to transport samples from a $q_{\bm{X}|\bm{Y}}$-weighted version of the simulation-based posterior $\hat{p}_{\bm{\Theta}|\bm{X}}(\bm{\theta}|\tilde{\bm{x}})$ toward the final corrected posterior $\hat{p}_{\bm{\Theta}|\bm{Y}}$. Note that both transports are required: ${T}_{\bm{X}}$ addresses the mismatches between simulated data and real observations, while $T_{\bm{\Theta}}$ refines parameter inference to align with the true posterior.
    }
    \label{fig:summary}
    %\vspace{-11pt}
\end{figure}

{In particular, we consider SBI applications where both $\bm{\theta}$ values and observations are quantities of physical meaning and interest, but for which it is impractical to model every detail of their generative process. This may generate model mispecification, a notion that needs to be tailored to our SBI setting.
The most standard definition of model misspecification in statistics, in both Bayesian \citep{walker_bayesian_2013,kelly_misspecification-robust_2024} and frequentist \citep{tomaselli2025robustsimulationbasedinference,park2025robust} contexts, corresponds to the so-called ${\cal M}$-open setting of \citet{bernardo1994bayesian}. A model is said to be misspecified  (${\cal M}$-open scenario) if there is no $\bm{\theta}$ value for which the assumed likelihood is able to reproduce the observed data. In this case, there is no such a thing  as a {\it true}  parameter $\bm{\theta}$. The inference target is then a so-called pseudo-true or projection parameter which minimizes a chosen discrepancy to the true data generating process, typically the Kullback-Leibler divergence in a maximum likekihood context \citep{tomaselli2025robustsimulationbasedinference, park2025robust,walker_bayesian_2013,kelly_misspecification-robust_2024}. Unfortunately, this value  is in general not  practically useful or desirable in  SBI contexts where $\bm{\theta}$ has a real-world meaning. In contrast, in the converse ${\cal M}$-closed  scenario, there exists a value of $\bm{\theta}$ for which the likelihood can reproduce the observed data but unfortunately, this property may not be a meaningful guarantee  in SBI.  As illustrated in 
  Section 2.2 and Appendix A.2 of \citet{wehenkel2025addressing} with a simple conjugate Gaussian example, the simulator may be well-specified according to the standard ${\cal M}$-closed  definition 
but still provides biased estimates of  the $\bm{\theta}$ of interest  when applied to real data.} {See Appendix \ref{app:missdef} for more details.}
{Other definitions of misspecification are more inclusive such as that of \citet{nott_bayesian_nodate} where a model is said to be misspecified if {\it we are not happy to act as if it is correct}. See also \citet{Burkner2023} for a more extensive discussion on what is {\it expected} from a well-specified model.
Effective approaches, particularly adapted to ABC settings,  to robustify inference include to exhibit summary statistics that mitigate the full data model misspecification \citep{frazier_2020b,huang_learning_2023,nott_bayesian_nodate,kelly_simulation-based_2025}. Typically insufficient summary statistics may be compatible with several models but designing such summaries is not the focus of our work. 
}

{We thus adapt the notion of misspecification to our SBI context. 
%We consider SBI applications where both $\bm{\theta}$ values and observations are quantities of physical meaning and interest, but for which it is impractical to model every detail of their generative process. 
Considering real or high-fidelity observations $\yv \in \mathbb{R}^d$, we  assume the existence of an unknown or ideal  data-generating process, with a likelihood $p_{\bm{Y}|\bm{\Theta}}$ and prior $p_{\bm{\Theta}}$ over parameters  of interest.}
SBI algorithms provide various approaches~\citep{papamakarios_fast_2016,lueckmann_flexible_2017,boelts_flexible_2022,pmlr-v119-hermans20a} to obtain approximate samples from the posterior {$\pthy \propto \pth \; p_{\bm{Y}|\bm{\Theta}}$} with the aid of deep generative models.  
In this paper, we focus on neural-based approaches that directly approximate the posterior distribution, often called neural posterior estimation (NPE). This corresponds to {the following form of posterior estimates, where the simulator's likelihood $p_{\bm{X}|\bm{\Theta}}$ {and an assumed prior $q_{\bm{\Theta}}(\bm{\theta})$} are used in place of the {target data likelihood $p_{\bm{Y}|\bm{\Theta}}$ and prior $p_{\bm{\Theta}}$,}
\begin{equation}
\hat{p}_{\bm{\Theta}|\bm{X}}(\bm{\theta}|\yv) \propto {q_{\bm{\Theta}}(\thetab)}\pxth(\yv | \thetab)~.
\end{equation}
{Following  \citet{schmitt_detecting_2024}, we use the word misspecification to refer to the {\it simulation gap} that may occur when the (simulator) operational generative model assumed during training differs from the (unknown) one inverted at test time. Misspecification may then arise from any of the assumed model components.
%, including nuisance or noise parameters omitted in the notation for simplicity. 
We will refer to a model as misspecified if any of the model components is undergoing an unacceptably large mismatch.}
Discrepancies between   $p_{\bm{Y}|\bm{\Theta}}$ and  $p_{\bm{X}|\bm{\Theta}}$, can severely degrade parameter inference using the above $\hat{p}_{\bm{\Theta}|\bm{X}}$. 
{Likewise, error in  prior modelling using $q_{\bm{\Theta}}$ instead of $\pth$ may bias posteriors, particularly in finite-sample settings.
%, while (iii) 
Unaccounted noise or contamination can  
lead to biased or overconfident posterior distributions, ultimately undermining inference reliability \citep{frazier_2020b,schmitt_detecting_2024}.}
%produce erroneous inference %\cite{schmitt_detecting_2024}}.
%In the Bayesian literature, this problem is referred to as \emph{model misspecification}~\citep{walker_bayesian_2013}.
{Misspecification generally reflects} limitations of the simulator or surrogate forward model—whether due to mathematical approximations, simplified dynamics, unmodeled noise, or insufficient computational resources to run more realistic simulations.
%Such mismatches can lead to biased or overconfident posterior distributions  \citep{frazier_2020b, schmitt_detecting_2024}, ultimately undermining inference reliability. 
In particular, modern SBI methods built on deep generative models are especially vulnerable to misspecification. They often perform poorly when faced with out-of-distribution data \citep{nalisnick2018do} and can fail dramatically across a wide range of real-world problems \citep{cannon_investigating_2022,schmitt_detecting_2024}.

A natural way to address imperfections in $\hat{p}_{\bm{\Theta}|\bm{X}}$ is to act to reduce its deviation from an estimator built on high-quality data.
{Following  recent literature \cite{wehenkel2025addressing, senouf2026inductive,krouglova2025multifidelitysimulationbasedinferencecomputationally}, we assume 
access to such accurate representations via a calibration dataset of parameter–observation pairs ${\cal D}_\text{cal}=\{(\rvtheta_j, \rvy_j)\}_{1\leq j \leq N_{\text{cal}}}$,
obtained either from high-fidelity simulations or from ground-truth controlled experiments. In experimental sciences,
experts typically collect such calibrated field data  to offset limits in models and sustain the development of  accurate digital twins.}
However, since such data may be scarce due to their potential prohibitive cost, the central challenge is to design strategies that maximally exploit the limited information they provide in order to correct $\hat{p}_{\bm{\Theta}|\bm{X}}$. 

 We propose a calibration method that leverages the flow matching paradigm of~\citet{lipman_flow_2023} to refine posterior estimators trained on simulations using only a small set of calibration samples linking parameters with real observations {or highly accurate simulations}. The principled ability of flows to efficiently model paths between distributions makes them good candidates to design corrective procedures agnostic to the type of misspecifications. Moreover, flows  have demonstrated high-scalability with state-of-the-art performance at large-scale image generation~\citep{Esser2024} and data-efficient performance in SBI~\citep{dax_flow_2023}.
% We have chosen such paradigm because of its recent demonstrations of both state-of-the-art performance at large-scale image generation~\citep{Esser2024} and data-efficient performance in SBI~  \citep{dax_flow_2023}.
Our approach proceeds in two stages. First, a posterior approximator $\hat{p}_{\bm{\Theta}|\bm{X}}$ is trained on abundant simulated data via NPE. Second, a reweighted version of this estimator is used to define a proposal distribution $\proposal$, which serves as the source in a flow matching model that learns a transport map to the true posterior $\pthy$. Crucially, this transport does not require explicit knowledge of the misspecification form: it aligns the simulation-based posterior with the true posterior supported by real observations. The resulting procedure enables accurate estimation of $\pthy$ despite the limited availability of calibration data. 
Figure~\ref{fig:summary} gives an overview of our method, which we call ``Flow Matching for Corrected Posterior Estimation`` (\ourmethod).
{Experiments on synthetic and real-world tasks show that \ourmethod\ is more robust to misspecification than standard SBI baselines, while being computationally efficient, amortized and applicable as a post-processing to any SBI model.}

The  paper is organized as follows. We begin with an overview of SBI under model misspecification and outline how it relates to our contribution. Next, we motivate and detail our methodology, introducing flow matching concepts and notation as needed. Finally, we present numerical experiments and conclude with a discussion of the results.

\section{Related work}
\label{sec:relatedwork}
Model misspecification in SBI has been studied both in the framework of Approximate Bayesian Computation (ABC)~\citep{frazier_2020b,bharti2022approximatebayesiancomputationdomain,pmlr-v130-fujisawa21a} and with modern neural-based approaches~\citep{kelly_misspecification-robust_2024, ward_robust_2022, huang_learning_2023, schmitt_detecting_2024}. These works focus on settings distinct from ours. For instance, \citet{ward_robust_2022} assume a known form of misspecification, while \citet{huang_learning_2023} penalize the NPE loss according to the distributional shift between summary statistics of $\xv$ and $\yv$. Other contributions, such as \citet{schmitt_detecting_2024}, restrict attention to detecting misspecification at inference time by imposing a Gaussian prior on the space of summary statistics.

An alternative perspective is to interpret misspecification as a noisy channel between simulated and real-world observations. In this view~\citep{wehenkel2025addressing, ward_robust_2022}, once simulated data $\xv$ are known, the real observation $\yv$ does not carry additional information about the parameters $\rvtheta$. Under this conditional independence hypothesis
\begin{equation}
\rvy \perp \rvtheta \mid \rvx~,
\label{eq:hyp}
\end{equation}
the true posterior distribution can be expressed as
\begin{equation}
\pthy(\bm{\theta} | \yv) = \int \pthx(\thetab | \xv) \pxy(\xv | \yv)  \, \text{d}\xv~,
\label{eq:posterior}
\end{equation}
which states that the true posterior distribution is a mixture of surrogate posteriors, weighted by the probability of each simulated outcome $\xv$ given $\yv$.
\citet{wehenkel2025addressing} exploit this idea by {assuming access to a ground truth calibration dataset 
%of parameter–observation pairs $\{(\rvtheta_i, \rvy_i)\}_{1\leq i \leq N_{\text{cal}}}$ 
%collected from costly or time-consuming experiments.
} Their method, named \ROPE, approximates $p_{\bm{\Theta}|\bm{Y}}$ by replacing $p_{\bm{X}|\bm{Y}}$ in Equation~(\ref{eq:posterior}) with a coupling $q_{\bm{X}|\bm{Y}}$ estimated via optimal transport~\citep{ot_peyre}, which links real observations $\yv$ and simulated data $\xv$. {A key limitation is that \ROPE\ requires access to a  full batch of test samples at inference time, preventing its use in online or sequential prediction scenarios.
This issue has been addressed by \citet{senouf2026inductive}. However, both approaches rely on} the conditional independence assumption (\ref{eq:hyp}), which does not always hold. For example, if model misspecification stems from the fact that the simulator neglects some parameters, for $\bm{\theta} = (\theta_1, \theta_2)$, $p_{\bm{X}|\bm{\Theta}}(\bm{x}|\bm{\theta}) = f(\theta_1)$ while $p_{\bm{Y}|\bm{\Theta}}(\bm{y}|\bm{\theta}) = g(\theta_1, \theta_2)$, then $\yv$ will still depend on the full parameter vector, violating conditional independence.

In contrast, the method \MFNPE~proposed by~\citet{krouglova2025multifidelitysimulationbasedinferencecomputationally} does not rely on conditional independence nor require access to the full test set. Their approach first trains a posterior approximator $p_{\text{LF}}(\rvtheta|\rvx)$ on low-fidelity simulations using standard NPE, then refines it with high-fidelity samples (i.e. coming from high-fidelity simulators or real data). 
{Similarly, our method does not assume (\ref{eq:hyp}) or any other conditional independence and does not make any specific assumptions on the misspecification form.} Our approach follows the same spirit of leveraging low-fidelity approximations obtained from a simulator, but uses them differently: the low-fidelity estimator plays a distinct role, and the subsequent corrections it undergoes are of a different nature, as detailed in the next section.

\section{Flow Matching Corrected Posterior Estimation}

%\textbf{Setup.}
We consider the general problem of sampling from a posterior distribution $p_{\bm{\Theta}|\bm{Y}}$ resulting from an \emph{unknown} likelihood $p_{\bm{Y}|\bm{\Theta}}$ over observation $\bm{y}\in \mathbb{R}^d$, {and prior} $p_{\bm{\Theta}}$ over a parameter vector $\bm{\theta}\in \mathbb{R}^p$. If the {model} were known, the posterior would be  available using the Bayes rule as $p_{\bm{\Theta}|\bm{Y}}(\bm{\theta} | \bm{y})\propto p_{\bm{Y}|\bm{\Theta}}(\bm{y} | \bm{\theta})p_{\bm{\Theta}}(\bm{\theta})$. Instead, we assume only access to  {a potentially approximate prior $q_{\bm{\Theta}}$ on $\bm{\theta}$ and to } an \emph{imperfect} stochastic simulator $S : (\rvtheta, \epsilon) \mapsto \xv$, generating low cost simulations $\xv\in \mathbb{R}^d$, with $\epsilon$  a random noise accounting for randomness in the generative process. 
%Faux en fait: {This includes $S(\rvtheta, \epsilon) = S(\rvtheta) +  \epsilon$ and thus Equation (\ref{eq:hyp}) as a particular case but is more general otherwise}. 
The simulator implicitly defines a generative model $p_{\bm{X}|\bm{\Theta}}$ whose posterior distribution 
%$p_{\bm{\Theta}|\bm{X}}(\bm{\theta}|\bm{x})$ 
can be approximated by an easy-to-sample model $\hat{p}_{\bm{\Theta}|\bm{X}}$,  using techniques from the SBI literature such as NPE. %\citep{greenberg_automatic_2019}. 
Unfortunately, since the simulator {or prior can be}  inaccurate, the SBI model $\hat{p}_{\bm{\Theta}|\bm{X}}$ is likely  not to provide an accurate approximation to the true posterior $p_{\bm{\Theta}|\bm{Y}}$,  in general,  no matter how accurately it approximates the simulator's posterior $p_{\bm{\Theta}|\bm{X}}$. However, one can reasonably expect such learned model to be informative about the true posterior. We propose to leverage the simulator $S$ and posterior model $\hat{p}_{\bm{\Theta}|\bm{X}}$ in addition to a small set of calibration pairs of high-quality data $\mathcal{D}_{\text{cal}} = \{(\bm{\theta}_j, \yv_j)\}_{1 \leq j \leq N_\text{cal}} $  from the joint distribution $p_{\bm{\Theta},\bm{Y}}$  to provide an accurate and efficient model $\hat{p}_{\bm{\Theta}|\bm{Y}}$ for the true posterior $p_{\bm{\Theta}|\bm{Y}}$.
 % \textcolor{green}{Such a dataset  typically represents scarce ground-truth measurements or high-fidelity simulations that are costly to obtain.}
%and is can be augmented by samples $\bm{x}_j$ using the simulator evaluated at parameter $\bm{\theta}_{j}$. 
Consequently, we assume that $N_\text{cal}$ is not large enough to provide an accurate posterior estimate from this dataset alone.  

We propose to use the flow matching paradigm to learn a dynamic transport map, a vector field, from a carefully designed source distribution $\pi_{\bm{\Theta}|\bm{Y}}$ towards the target posterior $p_{\bm{\Theta}|\bm{Y}}$ using the calibration dataset $\mathcal{D}_{\text{cal}}$. Increasing the proximity of the source and target distributions reduces the complexity of the flow and makes it easier to learn from a small number of samples. Intuitively,
smaller distributional gaps are likely to require fewer steps and improve sample complexity \citep{cui2024analysis,lin2025optimaltransportmodelalignedcoupling,kong2025compositeflowmatchingreinforcement,wang2025sourceguidedflowmatching}. Following this principle, we employ the simulator $S$ and the posterior approximation $\hat{p}_{\bm{\Theta}|\bm{X}}$, to design and train an informative source distribution that facilitates the learning of the vector field from few calibration samples. The resulting algorithm (Algorithm~\ref{alg:training} in Section \ref{sub:training}) combines simultaneous learning of both the source distribution and the vector field, ensuring the source is continually updated as the vector field becomes more accurate. Next, we detail the steps for learning the vector field given an informed source distribution, constructing the source itself and training them jointly.

\subsection{Data-efficient posterior flow matching}
%{Il y a des commandes pour les vectors fields et pdf dans commands.tex}
%Our goal is to estimate the posterior distribution $\pthy$ using a small dataset $\mathcal{D}_{\text{cal}}$ with high-fidelity calibration samples and an abundant dataset $\mathcal{D}_{\text{sim}}$ containing simulation samples. Denoting $p_{\bm{\Theta}}$ the prior over parameters, $p_{\bm{X}|\bm{\Theta}}$ the simulator likelihood, and $p_{\bm{Y}|\bm{\Theta}}$ the likelihood of real-word observations, we have
%\begin{equation}
%\begin{array}{rcl}
%\mathcal{D}_{\text{cal}} = \{(\bm{\theta}_j, \bm{x}_j, \bm{y}_j)\}_{j = 1}^{N_\text{cal}} &\sim& p_{\bm{\Theta}}(\bm{\theta})p_{\bm{X}|\bm{\Theta}}(\bm{x}|\bm{\theta}) p_{\bm{Y}|\bm{\Theta}}(\bm{y}|\bm{\theta}) \\[0.75em]\mathcal{D}_{\text{sim}} = \{(\bm{\theta}_i, \bm{x}_i)\}_{i = 1}^{N_\text{sim}} &\sim& p_{\bm{\Theta}}(\bm{\theta})p_{\bm{X}|\bm{\Theta}}(\bm{x}|\bm{\theta})~.
%\end{array}
%\end{equation}

%%
%\begin{eqnarray}
%\mathcal{D}_\text{cal} &=& \{(\bm{\theta}_j, \bm{x}_j, \bm{y}_j)\}_{j = 1}^{N_\text{cal}} \sim p_{\bm{\Theta}}(\theta)p_{\bm{X}|\bm{\Theta}}(\bm{x}|\bm{\theta}) p_{\bm{Y}|\bm{\Theta}}(\bm{y}|\bm{\theta})\\
%\mathcal{D}_\text{sim} &=& \{(\bm{\theta}_i, \bm{x}_i)\}_{i = 1}^{N_\text{sim}} \sim p_{\bm{\Theta}}(\theta)p_{\bm{X}|\bm{\Theta}}(\bm{x}|\bm{\theta})
%\end{eqnarray}
%with $N_{\text{cal}} \ll N_{\text{sim}}$. 
The calibration set $\mathcal{D}_\text{cal}$ is used to learn a transport map ${T_{\bm{\Theta}}:\mathbb{R}^{p} \times \mathbb{R}^d \to \mathbb{R}^{p}}$ from an easy-to-sample base (or source) distribution $\pi_{\bm{\Theta}}$ towards the posterior $\pthy$: 
\begin{equation}
    \bm{\theta}_0 \sim \pi_{\bm{\Theta}}(\bm{\theta}) \Rightarrow  \bm{\theta}_1 = T_{\bm{\Theta}}(\bm{\theta}_0, \bm{y}) \sim \pthyv~.
\end{equation}
%While a non-informative pre-determined proposal, such as a Gaussian distribution, is a sensible choice in general, the choice of the proposal is critical, in our setting, for learning an accurate transport map, due to limited access to calibration data. As we will discuss later, we will leverage simulation data to learn an informative proposal that simplifies learning of the transport map. 
To learn $T_{\bm{\Theta}}$, 
we use the flow matching framework of~\citet{lipman_flow_2023}, an approach that has proven effective for modeling complex target distributions, such as images \citep{Esser2024}, and has recently been applied successfully in simulation-based inference \citep{dax_flow_2023}. 
Flow matching consists of learning a time-dependent vector field $u_{\bm{\Theta}}$ capable of transporting a sample $\bm{\theta}$ from $\pi_{\bm{\Theta}}$ along a trajectory $(\bm{\theta}_{t})_{t\in [0,1]}$ starting from $\bm{\theta}_0=\bm{\theta}$, so that $\bm{\theta}_1$ is distributed according to  the target $\pthy$. The trajectory is obtained as a solution of the following ODE:  
\begin{equation*}
\frac{\dd\bm{\theta}_t}{\dd t}  = u_{\bm{\Theta}}\left(t, \bm{\theta}_t, \bm{y}\right),\quad \forall t\in [0,1],   \quad ~~\text{with}~~\bm{\theta}_0 = \bm{\theta}.
\end{equation*}
The time-dependent flow $\psi_{\bm\Theta}$ associated to the above ODE is simply given by samples along the path starting from the initial condition $\bm{\theta}$, i.e. $\psi_{\bm\Theta}(t,\bm{\theta},\bm{y}) := \bm{\theta}_t$. Such a flow allows defining the transport map from $\pi_{\bm{\Theta}}$ to $\pthy$ as $T_{\bm{\Theta}}(\bm{\theta}, \bm{y}):=\psi_{\bm{\Theta}}(1,\bm{\theta},\bm{y})$. 
The vector field is approximated with a deep neural network $\hat{u}_{\bm\Theta}$ trained on a \emph{guided} version of the conditional flow-matching loss from~\citet{lipman_flow_2023} with a linear interpolation strategy:
\begin{equation}
\begin{aligned}
  \mathcal{L}_{\bm{\Theta}}(\hat{u}_{\bm\Theta}) &= \mathbb{E}
\left[\int_0^1 \left\Vert \hat{u}_{\bm\Theta}(t, \bm{\theta}_t, \bm{y}) - (\bm{\theta}_1 - \bm{\theta}_0) \right\Vert^2 \mathrm{d}t\right],\\
{}& \bm{\theta}_t := (1-t)\bm{\theta}_0 + t \bm{\theta}_1 ~ \forall t \in [0,1]~,
\label{eq:theta_fm_loss}
\end{aligned}
\end{equation}
where the expectation is taken over  $(\bm{\theta}_1,\bm{y},\bm{\theta}_0) \sim p_{\bm{\Theta},\bm{Y}}(\bm{\theta}_1,\bm{y}) \pi_{\bm{\Theta}}(\bm{\theta}_0)$.  The calibration dataset $\mathcal{D}_{\text{cal}}$ can be used to provide joint samples $( \bm{\theta}_1,\bm{y})$ from  $p_{\bm{\Theta},\bm{Y}}(\bm{\theta}_1,\bm{y})$, thus providing an empirical version of the above objective. 
%Note that the samples $(\bm{y}, \bm{\theta}_1)$  are obtained using the fact that $p_{\bm{Y}}(\bm{y}) \pthyv = p_{\bm{Y}|\bm{\Theta}}(\bm{y} | \bm{\theta}) p_{\bm{\Theta}}(\bm{\theta})$ and the loss is approximated by its empirical version with high-fidelity samples $(\bm{\theta_j}, \bm{y}_j)$ from $\mathcal{D}_\text{cal}$. 

In most flow matching instances, the base distribution $\pi_{\bm{\Theta}}$ is set to a simple standardized Gaussian distribution, i.e. $\pi_{\bm{\Theta}}(\bm{\theta}) = \mathcal{N}(\bm{\theta} \mid \bm{0},\bm{I}_p)$. However, for settings such as the one we consider here, where $\mathcal{D}_\text{cal}$ is  small, this choice may result in very poor posterior approximators with high variance. Instead, we use 
the simulator 
%abundant low-fidelity simulation data in $\mathcal{D}_\text{sim}$ 
to construct a {more informative} data-driven source distribution $\pi_{\bm{
\Theta}} = \proposalsimp$ that acts as {a possibly low-quality surrogate of the true $p_{\bm{\Theta} | \bm{Y}}$.}
%an informed prior for inference. 
Training itself remains consistent with standard guided flow matching; the learned flow then serves to refine this source distribution using only the limited calibration data.

\subsection{Simulation-informed source distribution}\label{sec:proposal}

The source distribution $\proposalsimp$ plays a central role in our framework. Ideally, it should be very close to the true posterior distribution, so that the flow ${\psi}_{\bm{\Theta}}$ induces minimal corrections to $\bm{\theta}_0 \sim \pi_{\bm{\Theta}|\bm{Y}}$. Making use of the availability of the simulator $S$, a first natural possibility is to set
$\pi_{\bm{\Theta}|\bm{Y}}(\bm{\theta}|\bm{y}) = \hat{p}_{\bm{\Theta}|\bm{X}}(\bm{\theta}|\bm{y})$
where  
$\hat{p}_{\bm{\Theta}|\bm{X}}$ is a posterior approximation obtained via SBI. This  provides a reasonable approximation of the true $\pthy$ in the absence of misspecification but is likely to be a poor one in less favorable settings. 
 A natural alternative is to plug into $\hat{p}_{\bm{\Theta}|\bm{X}}$ a conditioning sample 
 from a distribution $q_{\bm{X}|\bm{Y}}(\bm{x}|\bm{y})$ with the same support as $p_{\bm{X}}(\bm{x})$ and informative of $\bm{y}$. This corresponds to considering a source distribution of the form 
 \begin{equation}
\label{eq:source}
    \pi_{\bm{\Theta}|\bm{Y}}(\bm{\theta}|\bm{y}) = \int \hat{p}_{\bm{\Theta}|\bm{X}}(\bm{\theta}|\bm{x}) \, q_{\bm{X}|\bm{Y}}(\bm{x}|\bm{y}) \, d \bm{x}.
\end{equation}
 The most straightforward choice for $q_{\bm{X}|\bm{Y}}$ is  $q_{\bm{X}|\bm{Y}}(\bm{x}|\bm{y}) := p_{\bm{X}|\bm{Y}}(\bm{x}|\bm{y})$, which would require, in addition to ${\cal D}_{\text{cal}}$, a large number of ground truth samples
  $(\bm{y},\bm{x})$ for training an approximator.
  Thus, we consider instead, 
   \begin{align}
    q_{\bm{X}|\bm{Y}}(\bm{x}|\bm{y}) &:=  \int p_{\bm{X}|\bm{\Theta}}(\bm{x}|\bm{\theta}) \, p_{\bm{\Theta} |\bm{Y}}(\bm{\theta}|\bm{y}) \, d\bm{\theta}, \label{def:qxy}
    \end{align}
that can be seen as a mixture and can be approximated using simulator $S$ and  joint calibration pairs $(\bm{\theta}_j,\bm{y}_j) \sim p_{\bm{\Theta}, \bm{Y}}(\bm{\theta}, \bm{y}) = p_{\bm{\Theta}|\bm{Y}}(\bm{\theta} | \bm{y})p_{\bm{Y}}(\bm{y})$. 
% For a given $\rvy_j \in {\cal D}_{\text{cal}}$,  we first make use of $\rvtheta_j$ as a sample from $p_{\bm{\Theta} |\bm{Y}}(\bm{\theta}|\bm{y}_j)$  and then sample  from $p_{\bm{X}|\bm{\Theta}}(\bm{x}|\bm{\theta}_j)$ or, equivalently, run the simulator $S$ from $\rvtheta_j$. This is typically used in lines 2 and 3 from Algorithm~(\ref{alg:sampler}).
Indeed, note that for each $\rvy_j \in {\cal D}_{\text{cal}}$, its associated $\rvtheta_j$ is a sample from $p_{\bm{\Theta} |\bm{Y}}(\bm{\theta}|\bm{y}_j)$ that can be plugged in the simulator to generate $\bm{x}_j \sim p_{\bm{X}|\bm{\Theta}}(\bm{x}|\bm{\theta}_j)$, leading to $\bm{x}_j \sim q_{\bm{X}|\bm{Y}}(\bm{x}|\bm{y_j})$. This is typically used in lines 1 and 2 of Algorithm~\ref{alg:sampler}.
{Note that $q_{\bm{X}|\bm{Y}} = p_{\bm{X}|\bm{Y}}$ if $\bm{x} \perp \bm{y} \; | \; \bm{\theta}$, which may not be true if $\rvy$ depends on the nuisance parameter $\epsilon$.}
%, as we discuss in more details in Section \ref{sub:training}. 

A sample $\rvtheta$ from the source distribution  (\ref{eq:source}), can be obtained   by first drawing a sample $\tilde{\bm{x}}$ from $q_{\bm{X}|\bm{Y}}(\tilde{\bm{x}}|\bm{y})$ in (\ref{def:qxy}), then setting $\bm{\theta}$ to a sample from $\hat{p}_{\bm{\Theta}|\bm{X}}(\rvtheta|\tilde{\bm{x}})$. As we can already easily sample from $\hat{p}_{\bm{\Theta}|\bm{X}}$, it only remains to provide a way to sample from $q_{\bm{X}|\bm{Y}}(\bm{x}|\bm{y})$ for any $\bm{y}$.  
To this end, we use the flow matching framework to construct a conditional transport map $T_{\bm{X}}:\mathbb{R}^{d} \times \mathbb{R}^{d} \to \mathbb{R}^d$, from a Gaussian distribution centered around $\bm{y}$ with isotropic variance $\sigma^2$, towards $q_{\bm{X}|\bm{Y}}(\bm{x}|\bm{y})$ i.e.: 
\begin{equation}
\label{eq:proposalflow}
   \!\! \bm{x}_0 \sim \mathcal{N}\!(\bm{x}|\bm{y}, \sigma^2\! \bm{I}_d) \to \tilde{\bm{x}} \!=\! T_{\bm{X}}(\bm{x}_0, \bm{y}) \sim q_{\bm{X}|\bm{Y}}(\bm{x}|\bm{y}).
\end{equation}
More specifically, we define $ T_{\bm{X}}(\bm{x}_0, \bm{y}) = {\psi}_{\bm{X}}(1, \bm{x}_0, \bm{y})$, where ${\psi}_{\bm{X}}$ is the flow associated to the velocity field neural approximator $\hat{u}_{\bm X}$ learned by minimizing the following loss: 
\begin{align}
%\begin{aligned}[t]
      \mathcal{L}_{\bm{X}}(\hat{u}_{\bm X}) &= \mathbb{E}\left[\!\int_0^1\! \!\!\left\Vert \hat{u}_{\bm X}(t, \bm{x}_t, \bm{y}) - (\bm{x}_1 - \bm{x}_0) \right\Vert^2\mathrm{d}t\right] \, \nonumber\\
      {}& ~\bm{x}_t := (1-t)\bm{x}_0 + \! t \bm{x}_1~ \forall t \in [0,1]\!
%\end{aligned}
\label{eq:x_fm_loss}
\end{align}
with 
 { $(\bm{y},  \bm{x}_1,\bm{x}_0)\!\sim\!q_{\bm{Y},\bm{X}}(\bm{y},\!\bm{x}_1\!) \mathcal{N}\!(\bm{x}_0|\bm{y},\sigma^2 \! \bm{I}_d)$}. The dataset $\mathcal{D}_{\text{cal}}$ and simulator $S$  can be used to provide joint samples $(\bm{y}, \bm{x}_1)$ from  $q_{\bm{Y},\bm{X}}(\bm{y},\bm{x}_1)$ as discussed later in Section \ref{sub:training}. We took inspiration from \citet{albergo2024stochasticinterpolantsdatadependentcouplings} to define a source distribution that induces a coupling between the base and target distributions through the conditioning variable, which greatly helps the training process under limited data. 

When the conditional independence hypothesis (\ref{eq:hyp}) is valid and choosing $q_{\bm{X}|\bm{Y}}(\bm{x}|\bm{y})= p_{\bm{X}|\bm{Y}}(\bm{x}|\bm{y})$, we can use Equation (\ref{eq:posterior}) to see that $p_{\bm{X}|\bm{Y}}(\bm{x}|\bm{y})$ is indeed the optimal choice for transporting data from $\bm{y}$-space to $\bm{x}$-space, since
{
\begin{gather*}
    \pthy(\bm{\theta}|\bm{y}) = \mathbb{E}_{\bm{X}|\bm{Y}}\left[p_{\bm{\Theta}|\bm{X}}\!(\bm{\theta}|\bm{x})|\bm{y}\right],~\\
 \tilde{\bm{{x}}} \sim p_{\bm{X}|\bm{Y}}(\bm{x}|\bm{y})~\text{and}~\bm{\theta} \sim p_{\bm{\Theta}|\bm{X}}\!(\bm{\theta}|\tilde{\bm{{x}}}) \!\Longrightarrow \!\bm{\theta} \sim \pthy(\bm{\theta}|\bm{y}).
\end{gather*}
}
In that case, learning only $\hat{u}_{\bm{X}}$ could be sufficient, since the proposal $\pi_{\bm{\Theta}|\bm{Y}}$ is already a good approximation to the true posterior. However, if the conditional independence is no longer valid, if $\hat{p}_{\bm{\Theta}|\bm{X}}$ is poorly trained, or $\hat{u}_{\bm{X}}$ is not optimal, the proposal would not have the flexibility to compensate for errors from each of its different parts. This is no longer an issue when using $\hat{u}_{\bm{X}}$ to define a source distribution for learning the vector field $\hat{u}_{\bm{\Theta}}$, as proposed in this work, since it does not rely on the validity of  Equation (\ref{eq:hyp}).

\subsection{Joint training of posterior and source distributions by flow matching}\label{sub:training}

We now face two optimization tasks: minimizing the loss in (\ref{eq:x_fm_loss}) to train the simulation-space vector field $\hat{u}_{\bm{X}}$
%, which defines the proposal $\pi=\proposal$, 
and minimizing (\ref{eq:theta_fm_loss}) to train the parameter-space field $\hat{u}_{\bm{\Theta}}$. Instead of solving them separately, we propose to optimize the following \emph{joint objective} for improved practical performance:
\begin{equation}
\begin{aligned}
\mathcal{L}_{\bm{X},\bm{\Theta}}\!(\!\hux\!,\!\huth\!)\!
= \!\mathbb{E}\!\Bigg[\!\!
    \int_0^1 \!\!\!
    &\big\Vert \huth\!(t, \!\bm{\theta}_t, \!\bm{y})\! -\! (\bm{\theta}_1 \!- \bm{\theta}_0)\!\big\Vert^2 \!\!\mathrm{d}t \\
    +\int_0^1 \!\!\!
    &\big\Vert \hux\!(t, \!\bm{x}_t, \!\bm{y}) \!- \!(\bm{x}_1 \!- \bm{x}_0)\!\big\Vert^2 \!\!\mathrm{d}t\!
\Bigg]\!
\end{aligned}
\end{equation}
\vspace{-.5cm}
\begin{align*}
\mbox{with } \; (\bm{\theta}_1, \bm{y}) 
&\sim p_{\bm{\Theta},\bm{Y}}, \\[0.3em]
(\bm{x}_0,\bm{x}_1,\bm{\theta}_0)
&\sim \mathcal{N}(\bm{x}_0;\bm{y}, \sigma^2 \bm{I}_d)q_{\bm{X}\mid\bm{Y}}(\bm{x}_1 \!\mid \bm{y})\pi_{\bm{\Theta}\mid\bm{Y}}(\bm{\theta}_0\! \mid \bm{y})
\end{align*}
 and  $\bm{\theta}_t$ and $\bm{x}_t$ are convex combinations as in Equations (\ref{eq:theta_fm_loss}) and (\ref{eq:x_fm_loss}). 
%A practical difficulty arises in the first term of the above objective: 
{In practice, at a given iteration $k$ of the optimization procedure, 
$(\boldsymbol{\theta}_1,\bm{y})$ are sampled uniformly from the calibration dataset, where $\boldsymbol{\theta}_1$ serves as target sample to the vector field $\hat{u}_{\bm{\Theta}}$. 
We then generate  $\bm{x}_{1}$ given $\bm{y}$ using (\ref{def:qxy}), that is  $\bm{x}_{1} \sim p_{\bm{X}|\bm{\Theta}}(\bm{x}|\bm{\theta}_1)$, or equivalently by setting $\bm{x}_{1}=S(\bm{\theta}_1, \epsilon)$ as explained in Section \ref{sec:proposal}. This $\bm{x}_{1}$ serves as target for $\hat{u}_{\bm{X}}$, while $\bm{x}_0$ given $\bm{y}$ is obtained from a Gaussian distribution centered at $\bm{y}$.
For the source distribution (\ref{eq:source}),
$\bm{\theta}_0$ is sampled from  $\hat{p}_{\bm{\Theta}|\bm{X}}(\bm{\theta}|\tilde{\bm{x}})$ where in contrast to $\bm{x}_1$, $\bm{\tilde{x}}$ is not sampled from the simulator $S$ but is generated using the ODE induced by  the current estimated vector field $\hux^{(k)}$ and starting from $\bm{x}_0$, $\tilde{\bm{x}} = \psi_{\bm{X}}^{(k)}(1,\bm{x}_0, \bm{y})$. This also makes $\bm{\theta}_0$  independent of $\bm{\theta}_1$ given $\bm{y}$. 
%This sampling procedure is summarized in Algorithm \ref{alg:sampler}. 
It follows that  compared to standard flow matching,  $\bm{\theta}_0$  is sampled from a source $\proposal$ which is evolving with $\hux^{(k)}$ during training.} Intuitively, by coupling the estimations of $\hux$ and $\huth$, this joint formulation forces $\huth$ to be robust to noisy or inaccurate samples from the source distribution: during early training stages, $\proposal$ may yield poor candidates $\bm{\theta}_0$, yet $\huth$ must learn to accommodate them. 
% {We hypothesize...}{In other words, the ideal source (\ref{eq:source}) is replaced by a mixture of approximations gradually approaching (\ref{eq:source})  as the distribution of 
% $\psi_{\bm{X}}^{(k)}(1,\bm{x}_0, \bm{y})$
% approaches (\ref{def:qxy}).  The mixture
This robustness is desirable at test time, where the source distribution may also generate imperfect samples for previously unseen observations $\bm{y}$. 
% The joint learning of the vector fields is summarized in Algorithm~(\ref{alg:training}).
% It depends on a sampling strategy (function \textsc{SampleTrainingTuple}) to generate a tuple $(\bm{y},\bm{\theta}_1,\bm{\theta}_0,\bm{x}_1,\bm{x}_0)$. Function \textsc{SampleTrainingTuple} is then detailed in Algorithm~\ref{alg:sampler} to specify how we handle the implicit dependence on $\hux$ to get the
% desired distribution. 
% Algorithm~\ref{alg:training} first  specifies how to train both velocity flows jointly by optimizing the objective $\mathcal{L}_{\bm{\Theta},\bm{X}}(\hat{u}_{\bm{X}},\hat{u}_{\bm{\Theta}})$ using the sampling procedure from Algorithm~\ref{alg:sampler}. 
% In practice,  time is sampled independently for the $\bm{X}$ and $\bm{\Theta}$ terms  (lines 5 and 6) to reduce bias. Then, because $\hux$ affects the distribution of $\rvtheta_0$ (via $\tilde{\rvx}$), the effective source distribution for $\huth$ is non-stationary; to mitigate instability we use small learning rates and gradient clipping to reduce training instability in the early stages. 

The joint learning of vector fields $\hat{u}_{\bm{X}}$ and $\hat{u}_{\bm{\Theta}}$ is summarized in Algorithm~\ref{alg:training}, which specifies how to optimize the objective $\mathcal{L}_{\bm{X},\bm{\Theta}}(\hat{u}_{\bm{X}},\hat{u}_{\bm{\Theta}})$ using the sampling procedure {described above and summarized in } Algorithm~\ref{alg:sampler} referred  to as function \textsc{SampleTrainingTuple}. In practice, variable $t$ is sampled i.i.d. for each of the terms $\bm{X}$ and $\bm{\Theta}$ from $\mathcal{L}_{\bm{X}, \bm{\Theta}}$ (lines 5 and 7) to reduce bias. Also, because $\hux$ affects the distribution of $\rvtheta_0$ (via $\tilde{\rvx}$), the effective source distribution for $\huth$ is non-stationary; to mitigate instability we use small learning rates and gradient clipping to reduce training instability in the early stages. 
Another crucial aspect in the above sampling procedure is to prevent gradients from propagating through the intermediate sample $\tilde{\bm{x}}$ when optimizing the vector field $\hat{u}_{\bm{X}}$ which would bias its training.

%%%%%% ICLR verison %%%%%%%
% \begin{algorithm}[hbtp]
% \caption{Joint flow training  for FMCPE}
% \label{alg:training}
% \begin{algorithmic}[1]
% \footnotesize
% \Require \textsc{SampleTrainingTuple} function,  %Calibration set $\mathcal{D}_{\mathrm{cal}}$, pretrained proposal $\hat{q}(\theta\mid x)$ (frozen), 
% trainable flows $\hux,\huth$ and minibatch size $B$.

% \Repeat
%     \State $\mathcal{L}\gets 0$
%     \For{$i=1$ {\bf to} $B$}
%         \State $(\rvy,\rvtheta_1,\rvtheta_0,\bm{x}_1,\bm{x}_0) \gets \textsc{SampleTrainingTuple}(\hux)$ \Comment{see Algorithm \ref{alg:sampler}}
%         \State Draw $t \sim \mathcal{U}[0,1]$ and set $\bm{x}_{t} = (1-t)\bm{x}_0 + t\,\bm{x}_1$
%         \State Draw \emph{another independent} $\tau \sim \mathcal{U}[0,1]$ and set $\boldsymbol{\theta}_{\tau} = (1-\tau)\boldsymbol{\theta}_0 + \tau\,\boldsymbol{\theta}_1$
%         \State $\ell_{\bm{X}} \gets \|\hux(t, \bm{x}_{t}, \bm{y}) - (\bm{x}_1 - \bm{x}_0)\|^2$
%          \State $\ell_{\bm{\Theta}} \gets \|\huth(\tau, \boldsymbol{\theta}_{\tau}, \bm{y}) - (\rvtheta_1 - \rvtheta_0)\|^2$
%         \State $\mathcal{L} \gets \mathcal{L} + (\ell_{\bm{X}} + \ell_{\bm{\Theta}} )$
%     \EndFor
%     \State $\mathcal{L} \gets \mathcal{L} / B$
%     \State Take one optimizer step on trainable parameters of $\hux$ and $\huth$ using $\mathcal{L}$
%     % \Comment{optionally: apply loss weighting, warm-up, or freeze $u_X$ for stability}
% \Until{convergence}
% \end{algorithmic}
% \end{algorithm}
%%%%%%%%%%%%%%%%%%%%%%%%%%%%%%%%%%%%

\begin{algorithm}[h!]
\footnotesize
\caption{Joint flow training for \ourmethod}
\label{alg:training}
\begin{algorithmic}[1]
\footnotesize
\REQUIRE \textsc{SampleTrainingTuple}; trainable flows $\hat{u}_{\bm{X}}$, $\hat{u}_{\bm{\Theta}}$; minibatch size $B$

\REPEAT
    \STATE $\mathcal{L} \gets 0$
    \FOR{$i = 1$ to $B$}
        \STATE $(\bm{y}, \bm{\theta}_1, \bm{\theta}_0, \bm{x}_1, \bm{x}_0)
        \gets \textsc{SampleTrainingTuple}(\hat{u}_{\bm{X}})$
        \COMMENT{see Algorithm~\ref{alg:sampler}}

        \STATE Draw $t \sim \mathcal{U}[0,1]$
        \STATE $\bm{x}_t \gets (1 - t)\bm{x}_0 + t\,\bm{x}_1$

        \STATE Draw \textit{another independent} $\tau \sim \mathcal{U}[0,1]$
        \STATE $\bm{\theta}_\tau \gets (1 - \tau)\bm{\theta}_0 + \tau\,\bm{\theta}_1$

        \STATE $\ell_{\bm{X}} \gets
        \|\hat{u}_{\bm{X}}(t,\bm{x}_t,\bm{y}) - (\bm{x}_1-\bm{x}_0)\|^2$

        \STATE $\ell_{\bm{\Theta}} \gets
        \|\hat{u}_{\bm{\Theta}}(\tau,\bm{\theta}_\tau,\bm{y})
        - (\bm{\theta}_1-\bm{\theta}_0)\|^2$

        \STATE $\mathcal{L} \gets \mathcal{L} + \ell_{\bm{X}} + \ell_{\bm{\Theta}}$
    \ENDFOR

    \STATE $\mathcal{L} \gets \mathcal{L} / B$
    \STATE Update parameters of $\hat{u}_{\bm{X}}$ and $\hat{u}_{\bm{\Theta}}$
\UNTIL{convergence}
\end{algorithmic}
\end{algorithm}

\begin{algorithm}[h!]
\footnotesize
\caption{\textsc{SampleTrainingTuple}$(\hat{u}_{\bm{X}})$}
\label{alg:sampler}
\begin{algorithmic}[1]
\footnotesize
\REQUIRE Calibration set $\mathcal{D}_{\mathrm{cal}}$; simulator $S$;
pretrained $\hat{p}_{\bm{\Theta}|\bm{X}}(\bm{\theta}\mid\tilde{\bm{x}})$ (frozen);
velocity field $\hat{u}_{\bm{X}}$

\STATE Sample $(\bm{\theta}_1,\bm{y}) \sim \mathcal{D}_{\mathrm{cal}}$
\COMMENT{calibration sample}

\STATE Sample $\bm{x}_1$ using simulator $S$ evaluated at $\boldsymbol{\theta}_1$
\COMMENT{see Equation~(\ref{def:qxy})}

\STATE Draw base sample $\bm{x}_0 \sim \mathcal{N}(\bm{x}_0, \bm{y},\sigma^2 I)$
\COMMENT{base distribution}

\STATE Solve ODE
$\frac{d\bar{\bm{x}}_t}{dt} = \hat{u}_{\bm{X}}(t,\bar{\bm{x}}_t,\bm{y})$
with $\bar{\bm{x}}_0=\bm{x}_0$

\STATE $\tilde{\bm{x}} \gets T_{\bm{X}}(\bm{x}_0;\bm{y}) := \bar{\bm{x}}_1$
\COMMENT{current flow map}

\STATE $\tilde{\bm{x}} \gets \text{StopGradient}(\tilde{\bm{x}})$

\STATE Sample $\bm{\theta}_0 \sim
\hat{p}_{\bm{\Theta}|\bm{X}}(\bm{\theta}\mid\tilde{\bm{x}})$

\STATE \textbf{return} $(\bm{y},\bm{\theta}_1,\bm{\theta}_0,\bm{x}_1,\bm{x}_0)$
\end{algorithmic}
\end{algorithm}

% \vspace{-5pt}

\section{Experiments}

\begin{figure*}[hbtp]
    \centering
    \includegraphics[width=0.85\linewidth]{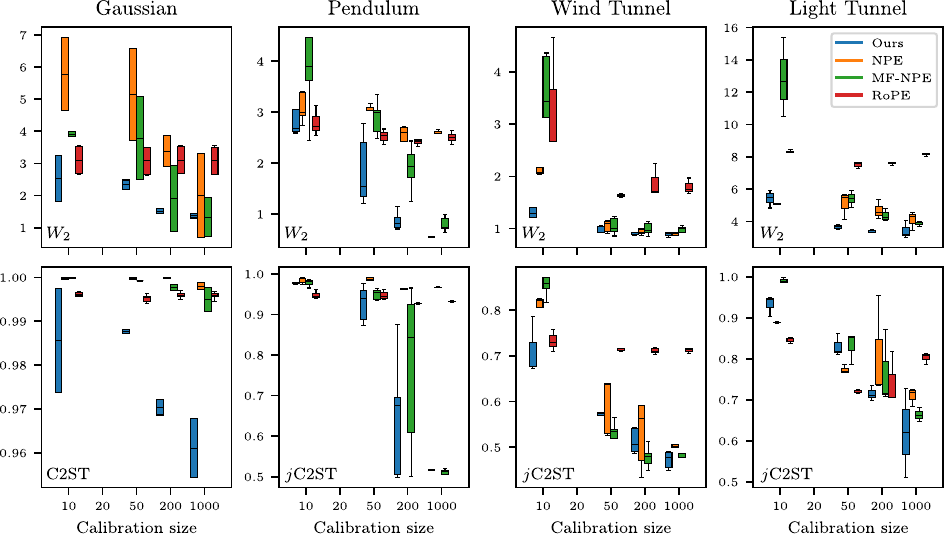}
    \caption{\footnotesize Wasserstein distance (top row, $\boldsymbol{\downarrow}$ is better) and C2ST/$j$C2ST (bottom row, $\boldsymbol{\downarrow}$ is better) with respect to an increasing calibration set size $N_{\text{cal}}\in \{10,50,200,1000\}$. Each boxplot shows the distribution of metric values across five independent runs, each using a different randomly chosen calibration set. Note that for the Gaussian Task (left), the C2ST and $W_2$ distance are averaged over 3 different observations chosen at random.
    }
    \label{fig:boxplots}
\end{figure*}

We benchmark our method against three baselines, NPE, \MFNPE, and \ROPE. The comparison is carried out on four tasks described in Section \ref{sec:tasks}, using various evaluation metrics specified in Section \ref{sec:metrics}. We follow the implementation from~\cite{wehenkel2025addressing} for NPE and train it using only the limited calibration data $\mathcal{D}_{\text{cal}}$. Note that since the simulator is not used, we expect this baseline NPE to fail as the complexity of the generating process increases. For each experiment, all evaluations are performed on a test set $\mathcal{D}_{\text{test}} = \{(\bm{\theta}_j, \bm{y}_j)\}_{1\leq j \leq N_\text{test}}$ with $N_{\text{test}} = 2000$ unless otherwise specified, and each metric is reported for different sizes $N_{\text{cal}}$ of the calibration set. Calibration sets are constructed in an expanding manner, gradually adding new samples to an initial set, to limit the sources of variability in the comparison; see Appendix \ref{app:exp_setup} for details.

All experiments described below are implemented in Python using \texttt{pytorch}~\citep{pytorch} and \texttt{mlxp}~\citep{mlxp}. We also use \texttt{nflows}~\citep{nflows} and the \texttt{dingo}~\citep{Dax:2021tsq} packages for the implementation of  normalizing flows.
{Our code is available at \url{https://github.com/pruhlmann/FMCPE}.}

\subsection{Tasks}
\label{sec:tasks}

{We first evaluate the methods ability to handle simulator misspecification.}
Our experiments consist of two synthetic and two real-world tasks. Setup and implementation details can be found in Appendix \ref{app:exp_setup}. 

$\blacktriangleright$ \TaskA:~A multivariate Gaussian model is considered, with $\rvtheta \in \mathbb{R}^3$ and $\rvtheta \sim \mathcal{N}(\mu_{\theta}, \Sigma_{\theta})$. Both $\rvx \in \mathbb{R}^{10}$ and $\rvy \in \mathbb{R}^{10}$ follow multivariate Gaussian distributions centered on different linear combinations of $\rvtheta$. 

% \textbf{Task B (Altered Gaussian) :} We introduce nonlinearities in the simulator through the sinh-arcsinh normal distribution \citep{sinh}. The prior is again a multivariate Gaussian and misspecification is introduced through a linear transformation of the output $\rvy = A\rvx + b + \varepsilon$ with $\varepsilon \sim \mathcal{N}(0,\sigma_{\text{noise}})$.

$\blacktriangleright$ \TaskB:~The damped pendulum \citep{PhysVae} models the oscillations of a mass around a fixed attachment point. Parameters $\rvtheta = [A, \omega_0]$ are the oscillation amplitude $A$ and the natural frequency $\omega_0$. Simulations are generated from a simplified model that omits friction forces, thus creating a systematic misspecification relative to the real dynamics. Both observations and simulations are real-valued time-series of length $200$.
 
$\blacktriangleright$ \TaskC:~This task from~\citet{Gamella2025} consists of measuring the air pressure inside an horizontal tube where air is being pushed through by two controllable fans at both ends. 
%The parameter of interest is the opening angle (in degrees) of a hatch on the side. We aim 
The goal is to infer the opening angle  (in degrees) of a hatch $H \in [0,45]$  on the side, given the pressure values inside the tunnel after applying a short power impulse to the intake fan. For the simulator, we use the model A2C3 from \citep[Appendix IV]{Gamella2025}.

$\blacktriangleright$ \TaskD:~In this task from~\citet{Gamella2025}, a camera is capturing light passing through two linear polarizers inside an elongated chamber. The goal is to predict RGB values of the light source and  the polarizer effect $\alpha \in [0,1]$, which is a function of the polarizer angle, so that $\rvtheta \!:= \![R,G,B,\alpha] \!\in \![0,255]^3 \!\times \![0,1]$. The simulator is a simplification of the real world process and described in  \citep[Appendix IV]{Gamella2025} (Model F1). Observations are RGB images of size $(W,H,C)=(64,64,3)$ produced by either the simulator or the real apparatus.  Model misspecification arises because the simulator omits certain physical effects present in the real measurements.

{$\blacktriangleright$ Prior misspecification: We then also illustrate the impact of prior misspecification using two of the previous tasks,  modifying the priors in the \TaskA\ and \TaskC\ examples. Details are in Appendix \ref{app:prior_misspec}.}

\subsection{Evaluation Metrics {and Calibration Diagnostics}}
\label{sec:metrics}
 {${\cal D}_{\text{test}}$ generally provides a single ground truth value $\rvtheta_j$ for each  $\rvy_j$,  so that metrics cannot be computed on parameter samples only as this would require multiple samples from the true posterior $p(\rvtheta|\rvy_j)$. Instead, for each $\rvy_j$ we generate one $\tilde{\rvtheta}_j$ from each learned posterior and  compare  ${\cal D}_{\text{test}}$ to the set of pairs $\{(\tilde{\rvtheta}_j, \bm{y}_j)\}_{1 \leq j\leq N_\text{test}}$ as samples from a joint distribution, except for  the \TaskA\ task since the ground truth posterior distribution is available in this case.}
We assess the performance of our method performance based on the following four metrics. 

\textbf{Classifier Two-Sample Test (C2ST) and joint C2ST ($j$C2ST)}. The C2ST \citep{lopez-paz2017revisiting} measures the discrepancy between two distributions by training a binary classifier to distinguish samples drawn from them. The test statistic is the classifier’s accuracy on samples from a test set, which is equal to chance-level (usually 0.5) for indistinguishable distributions and approaches $1$ as the distributions diverge. 
{The metric on parameter values used for the \TaskA~example is then denoted by C2ST while for all other tasks we write $j$C2ST to indicate that the metric is computed over joint samples.}
%This is the metric used for the \TaskA~example. For the other tasks, since our reference ${\cal D}_{\text{test}}$ provides a single ground truth value $\rvtheta_j$ for each  $\rvy_j$,
%standard C2ST cannot be used, as it would require multiple samples from the true posterior $p(\rvtheta|\rvy_j)$. Instead, for each $\rvy_j$ we generate one $\tilde{\rvtheta}_j$ from the learned posterior and  compare  ${\cal D}_{\text{test}}$ to the set of pairs $\{(\tilde{\rvtheta}_j, \bm{y}_j)\}_{1 \leq j\leq N_\text{test}}$ as samples from a joint distribution, justifying the name $j$C2ST. 
 
\textbf{Wasserstein Distance} ($W_2$). To provide a geometry-aware notion of distributional similarity, the  Wasserstein distance for the L$_2$ cost is computed between samples  from the approximate and true joint distribution $p(\rvtheta,\rvy)$, 
{except for the \TaskA\ task, unless otherwise specified.}
\vspace{-.2cm}
\begin{align*}
\begin{aligned}
W_2 =&\min_{\gamma \in \mathbb{R}_+^{N_{\text{test}} \times N_{\text{test}}}} \left( \sum_{i,j}\gamma_{i,j}\|(\rvtheta_i, \rvy_i)-(\tilde{\rvtheta}_j,\rvy_j)\|^2_2\right)^\frac{1}{2}~\\
&\text{such that}~\gamma .\mathbf{1} = \tfrac{1}{N_{\text{test}}}~\text{and}~\gamma^T.\mathbf{1}= \tfrac{1}{N_{\text{test}}} \; .
\end{aligned}
\end{align*}
%For the Gaussian task, the $W_2$ distance is computed between parameter samples only since the ground truth posterior distribution is available.

\textbf{Maximum Mean Discrepancy} (MMD).  
We compute the Maximum Mean Discrepancy between samples from the true and approximate joint distributions using a radial basis function (RBF) kernel. Given samples $\{(\rvtheta_i,\rvy_i)\}_{1\leq i \leq N_{\text{test}}}$ and $\{(\tilde{\rvtheta}_j,\rvy_j)\}_{1\leq j \leq N_{\text{test}}}$, the squared MMD is
\begin{align*}
\begin{aligned}
\mathrm{MMD}^2
&= \!\frac{1}{N_{\text{test}}(N_{\text{test}} -1)}\! \sum_{i \neq i'} k_{\sigma}\!\left((\rvtheta_i,\rvy_i),(\rvtheta_{i'},\rvy_{i'})\right)\\ 
&\quad + \!\frac{1}{N_{\text{test}}(N_{\text{test}} -1)}\! \sum_{j \neq j'} \!k_{\sigma}\!\left((\tilde{\rvtheta}_j,\rvy_j),(\tilde{\rvtheta}_{j'},\rvy_{j'})\right) \\
&\quad - \frac{2}{N_{\text{test}}^2} \sum_{i,j} k_{\sigma}\!\left((\rvtheta_i,\rvy_i),(\tilde{\rvtheta}_j,\rvy_j)\right),
\end{aligned}
\end{align*}
with $k_{\sigma}(x,x') = \exp\!\left(-\frac{\|x - x'\|_2^2}{2\sigma^2}\right)$ and $\sigma=10$ in our experiments.

\textbf{Mean Squared Error (MSE)}. The average mean squared error between $M$ generated samples $\{\tilde
{\rvtheta}_j^i\}_{1\leq i \leq M}$ and the ground truth parameters $\{\rvtheta_j\}_{1\leq j \leq N_{\text{test}}}$ for each observation $\rvy_j$ in $\mathcal{D}_{\text{test}}$, is given by 
\begin{equation*}
    \textrm{MSE} = \dfrac{1}{N_{\text{test}}}\dfrac{1}{M}\sum_j^{N_{\text{test}}} \sum_i^{M} \| \tilde{\rvtheta}^i_j - \rvtheta_j \|^2_2 \; .
\end{equation*}
MSE is a good accuracy measure for unimodal posteriors, which is often the case in our experiments.

{\textbf{Posterior calibration diagnostics (SBC).} 
We  also report the simulation-based calibration (SBC) diagnostic of \citet{talts2020validating}. For each 1D marginals of the posterior, SBC provides an histogram whose {deviation from uniformity} is indicative of a potential issue in the posterior, possibly too narrow, too wide or skewed. 
%Unfortunately, SBC does not provide sufficient conditions for posterior correctness.
{ In contrast, the absence of a significant deviation offers no guarantee that the posterior is correct, as uniformity trivially occurs when the posterior and the prior are the same.} SBC requires $N$  samples from the true generative process $\big\lbrace (\rvtheta_j,\rvy_j)\big\rbrace_{1 \leq j \leq N}$ and $L$ independent draws from the estimated posteriors for each $\rvy_j$. The procedure is detailed in Appendix \ref{app:sbc}.
}
{In practice, the sensitivity of SBC may be limited as true samples may only be available from a  calibration data set of small size. }

\subsection{Results and discussion}
\label{sec:results}

{Our analysis is detailed for the four tasks exhibiting simulator misspecifications. Additional experiments illustrating misspecifications in prior distributions are reported in Appendix \ref{app:prior_misspec} with similar conclusions.}
Figure~\ref{fig:boxplots} and Appendix Figure \ref{fig:all_metrics_all_methods}, first illustrate the impact of  $N_{\text{cal}}$ on posterior estimation, {for the compared methods and various metrics}. 
All metrics decrease as $N_{\text{cal}}$ increases, highlighting the importance of having enough calibration data to correct misspecification. 
% Note that for \TaskC~we sometimes observe $j$C2ST values below 0.5. This is due to an insufficient test set with respect to the dimensions of the problem, causing the classifier used for $j$C2ST to not be sufficiently well trained ($N_{\text{test}}$ was set to 5000 for this task, the maximal number of validation data points provided by the \texttt{causal-chamber} package).

 For a given ground truth couple $(\rvtheta^*, \yv^*) \in {\cal D}_{\text{test}}$, Figure \ref{fig:kde_plot} illustrates the posterior density of the first two components of $\rvtheta$ for each method  for the \TaskA~and~\TaskB\ tasks, 
 except \ROPE\ whose too flat posterior estimate hinders the other posterior distributions.
 Similar plots for the other tasks are provided in the Appendix in Figures \ref{fig:kde_plot_wind_tunnel} (\TaskC) and~\ref{fig:kde_plot_light_tunnel} (\TaskD).

% \begin{figure}
%     \centering
%     \includegraphics[width=\linewidth]{figures/kde_joint_obs3_seed6.pdf}
%     \caption{Caption}
%     \label{fig:placeholder}
% \end{figure}
\begin{figure*}[h!]
    \centering
% First subfigure
    \begin{subfigure}[b]{0.36\linewidth}
        \centering
         \includegraphics[width=\linewidth]{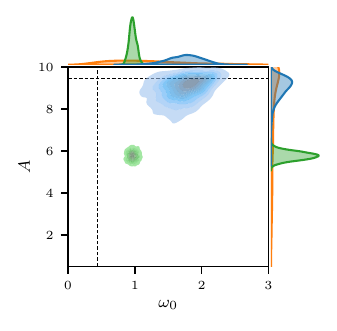}
        %\includegraphics[width=\linewidth]{figures/kde_joint_obs0_seed8_ncal10.pdf}
        % \caption{$N_{\text{cal}}=10$}
        \label{subfig:kde_plot_pendulum_10}
    \end{subfigure}
    \hspace{-12pt}
    % Second subfigure
    \begin{subfigure}[b]{0.31\linewidth}
        \centering
        \includegraphics[width=\linewidth]{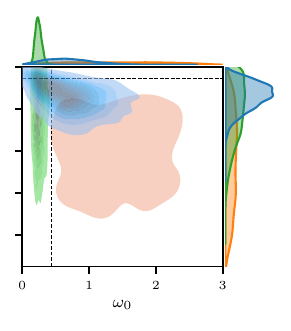}
      %  \includegraphics[width=\linewidth]{figures/kde_joint_obs0_seed8_ncal50.pdf}
        % \caption{$N_{\text{cal}}=50$}
        \label{subfig:kde_plot_pendulum_50}
    \end{subfigure}
    \hspace{-12pt}
    % Third subfigure
    % \vspace*{-1.0cm}
    \begin{subfigure}[b]{0.31\linewidth}
        \centering
        \includegraphics[width=\linewidth]{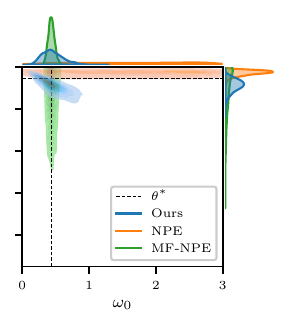}
       % \includegraphics[width=\linewidth]{figures/kde_joint_obs0_seed8_ncal200.pdf}
        % \caption{$N_{\text{cal}}=200$}
        \label{subfig:kde_plot_pendulum_200}
    \end{subfigure}

\begin{subfigure}[b]{0.38\linewidth}
        \centering
        \includegraphics[width=\linewidth]{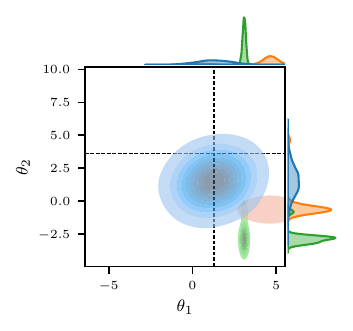}
        \caption{$N_{\text{cal}}=10$}
        \label{subfig:kde_plot_high_dim_gaussian_10}
    \end{subfigure}
    \hspace{-11pt}
    % Second subfigure
    \begin{subfigure}[b]{0.315\linewidth}
        \centering
        \includegraphics[width=\linewidth]{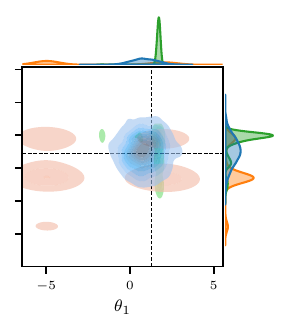}
        \caption{$N_{\text{cal}}=50$}
        \label{subfig:kde_plot_high_dim_gaussian_50}
    \end{subfigure}
    \hspace{-11pt}
    % Third subfigure
    \begin{subfigure}[b]{0.315\linewidth}
        \centering
        \includegraphics[width=\linewidth]{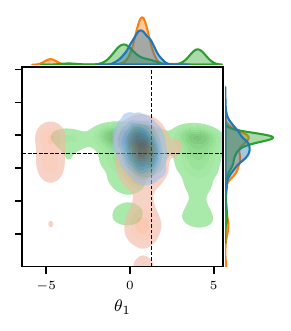}
        \caption{$N_{\text{cal}}=200$}
        \label{subfig:kde_plot_high_dim_gaussian_200}
    \end{subfigure}
\caption{\footnotesize  Kernel density estimates of joint and marginal samples for \TaskB~(first row) and  \TaskA~(second row). For a given  $\bm{y}^* \!\!\in {\cal D}_{\text{test}}$, we draw $\{\tilde{\bm{\theta}}_i\}_{1\leq i \leq 2000}$, for each method and 3 calibration sizes $N_{\text{cal}} \!\in\!\{ 10,50,200\}$. Dotted black lines indicate the true parameter $\rvtheta^*$ that generated $\rvy^*$.}
    \label{fig:kde_plot}
\vspace{-5pt}
\end{figure*}

%\paragraph{Posterior calibration.}

{SBC diagnostics  are reported in Appendix \ref{app:sbc}, Figures \ref{fig:sbc_gaussian} and \ref{fig:sbc_wt}, for our method and the other baselines. }

\paragraph{Comparison to baselines.}
All experiments demonstrate that our method consistently outperforms all baselines, quantitatively and qualitatively. In Figure~\ref{fig:boxplots}, we observe that for the \TaskA~task, our approach achieves better performance than plain NPE, fine-tuned \MFNPE, and \ROPE. This gap becomes more striking in Figure~\ref{fig:kde_plot} (second row), where both NPE and \MFNPE~produce multimodal posteriors even when the true posterior is unimodal. In contrast, our method yields unimodal posteriors that are better calibrated and centered on the ground-truth parameter $\bm{\theta}^*$, and this even  for a  low calibration size.
For the \TaskB~and \TaskD\ tasks, NPE fails to capture the intricate dependencies between parameters and observations—even when $N_{\text{cal}} = 1000$. In these cases, our method achieves superior performance in both $j$C2ST and $W_2$, while also exhibiting lower variance across draws of the calibration data set compared to \MFNPE~(Figure \ref{fig:boxplots}). Figure~\ref{fig:kde_plot} further illustrates that \MFNPE~often produces overly sharp posterior distributions in one or more dimensions, yet fails to recover the true parameter $\bm{\theta}^*$. In contrast, our method consistently recovers $\bm{\theta}^*$ with higher accuracy.
We hypothesize that this behavior stems from the \MFNPE~training procedure.  \MFNPE~learns a neural encoder $h_{\omega}(\bm{x})$ to extract latent representations from simulated data $\bm{x}$. During fine-tuning, however, this encoder is evaluated on real observations $\bm{y}$, whose distribution differs from that of $\bm{x}$. This distributional shift leads to erroneous latent representations $h_{\omega}(\bm{y})$, and consequently to biased posteriors. While this issue diminishes as $N_{\text{cal}}$ increases, it remains pronounced at small calibration sizes ($N_{\text{cal}}=10$ or $50$).
The \TaskB~task provides further evidence supporting this analysis. The misspecification arises from exponential damping, which predominantly affects the estimation of the oscillation amplitude $A$, as seen in Figure \ref{fig:kde_plot} (first row). In contrast, the estimation of the frequency $\omega_0$ remains accurate, since it is less sensitive to the damping mismatch.
In the real-world tasks (\TaskC~and \TaskD), our approach yields substantially better results at small calibration sizes, particularly in terms of $W_2$, and remains competitive or superior for larger calibration sets. Appendix~\ref{app:plots_metrics} provides additional visualizations of the posterior estimates for these tasks. {The impact of our correction is also illustrated in 
Figures \ref{fig:before_after_gaussian} to \ref{fig:before_after_lt},
%\ref{fig:before_after_pendulum},\ref{fig:before_after_wt}  and %\ref{fig:before_after_lt}, 
comparing  the posteriors before and after calibration.}

{SBC results show that our method is better calibrated as $\ncal$ grows and outperforms  baselines such as NPE and \MFNPE . For the \TaskA~ task, \ROPE\ shows less deviation from uniformity at low $\ncal$  but with no significant improvement when $\ncal$ increases. We believe this is due to the regularization introduced by parameters $\gamma$ and $\rho$ in the OT formulation,
which bias the \ROPE\ posterior toward the prior.}

\paragraph{Ablation study.}

To provide more insight on the different \ourmethod\ components, we also discuss in Appendix~\ref{app:plots_metrics} the  choice of using a joint optimization scheme and compare \ourmethod~to other alternatives where we isolate each component of the method to ponder its effect on the final results. Figure~\ref{fig:ablation} shows that \ourmethod~achieves better or comparable performances than the other methods being tested, in particular, it yields better results  than training both flows in a sequential manner. 

{For completeness, we also conducted an experiment with a  Gaussian model to check the effect of the conditional independence hypothesis (\ref{eq:hyp}) when we only use the mapping $T_{\bm{X}}$. Experiment details are reported in Appendix \ref{app:hyp_test}. Results in Table \ref{tab:ablation_flow_theta} show that when the conditional hypothesis holds, using only $T_{\bm{X}}$ is sufficient to recover the ground truth posterior, but as more and more dependence between $\rvtheta$ and $\rvy$ is introduced in the model, results start to degrade.}

%\paragraph{Posterior calibration.}

%In addition to to the metrics above, we reported the expected coverage of our methods and the other baselines Figures XXX (see Appendix \ref{app:sbc}).

%We also performed Simulation-Based Calibration (SBC) 

\paragraph{Training and Inference times.}

Times for \ourmethod\ and the baselines in Figure \ref{fig:boxplots} are reported in Appendix \ref{app:timing}, Tables  \ref{tab:training_rowstack} and  \ref{tab:inference_rowstack}. \ourmethod\ is more expensive to train but remains practical, requiring only 5–6 minutes per task, with  space for speed up, as our current implementation does not optimize the flow sampling step. At inference time, \ourmethod\ is faster than \ROPE\ for all tasks and calibration sizes.

%\sout{and illustrated in Figure \ref{fig:ablation} an ablation study investigating how different aspects of our method contribute to its final performance. We have considered a sequential training scheme where the proposal map $T_{\bm{\Theta}}$ is learned first and the calibration map $T_{\bm{X}}$ is fitted afterwards; an NPE+proposal approach where the proposal is trained only on calibration data and then fed directly into an NPE (analogous in spirit to the Robust NPE (RNPE) procedure from \citep{ward_robust_2022} but with a learned proposal instead of MCMC–based corrections); and a proposal+NPE variant where the proposal is trained using the likelihood estimator evaluated on $\bm{y}$-data. We observe the importance of the joint training of \ourmethod~as well as the use of a carefully chosen data-driven source distribution.}

\section{Conclusion}

We tackled model misspecification in SBI by combining scarce calibration data with abundant simulations. Our method builds a proposal posterior from both sources and refines it via flow matching, producing posterior estimates that are more accurate and better calibrated than standard SBI baselines. Importantly, our proposal is also computationally efficient, as it can leverage off-the-shelf SBI posterior distributions as proposals, requiring only lightweight refinement with calibration data.

{To complement these experimental observations, It would be interesting to further investigate whether useful bounds could be derived to quantify the quality of our proposed correction and to possibly derive  practical guidelines to set an effective calibration set size with respect to the task dimension and misspecification. }

%procedure could be an effective calibration set size could more precisely quantified with respect to the task dimension and misspecfication, in order to provide practical guidelines. Looking forward, our framework opens the door to further extensions, such as exploring richer proposal architectures, integrating domain adaptation techniques into the mapping between simulated and real data, or applying the method to large-scale scientific simulators where model misspecification is unavoidable.

While severe misspecifications in high dimensions may still require larger calibration sets, our results show that even small amounts of real data can substantially improve inference quality. This highlights the promise of our framework as a practical and scalable way to bring SBI closer to real-world scientific applications, and opens exciting opportunities for richer proposal architectures, domain adaptation techniques, and deployment on large-scale simulators where misspecification is inevitable.

\clearpage 

\section*{Impact Statement}
This paper presents a methodological contribution to simulation-based inference. % Its primary impact is expected to be within the research community, by providing tools to study and mitigate model misspecification when learning posterior distributions from simulations and limited real data
Our method is not tied to any specific application domain and does not introduce new capabilities that would raise ethical or societal concerns beyond those already present in machine learning. We therefore do not anticipate significant negative societal impacts arising directly from this work.
%{Future work would include  quantifying the calibration set size required
%for a desired accuracy, with respect to the task dimension and misspecification.}

%\subsubsection*{Reproducibility Statement}

 \section*{Acknowledgments}
{ The authors would like to thank Antoine Wehenkel for his insights and helpful remarks on the \ROPE\ algorithm. PLCR was supported by a national grant managed by the French National Research Agency (Agence Nationale de la Recherche) attributed to the SBI4C project of the MIAI AI Cluster, under the reference ANR-23-IACL-0006. MA was supported by the ANR project BONSAI (grant ANR-23-CE23-0012-01).}

\bibliography{icml2026}
\bibliographystyle{icml2026}
% this must go after the closing bracket ] following \twocolumn[ ...

%%%%%%%%%%%%%%%%%%%%%%%%%%%%%%%%%%%%%%%%%%%%%%%%%%%%%%%%%%%%%%%%%%%%%%%%%%%%%%%
%%%%%%%%%%%%%%%%%%%%%%%%%%%%%%%%%%%%%%%%%%%%%%%%%%%%%%%%%%%%%%%%%%%%%%%%%%%%%%%
% APPENDIX
%%%%%%%%%%%%%%%%%%%%%%%%%%%%%%%%%%%%%%%%%%%%%%%%%%%%%%%%%%%%%%%%%%%%%%%%%%%%%%%
%%%%%%%%%%%%%%%%%%%%%%%%%%%%%%%%%%%%%%%%%%%%%%%%%%%%%%%%%%%%%%%%%%%%%%%%%%%%%%%
\newpage
\appendix
\onecolumn

\section{Misspecification in SBI}
\label{app:missdef}

{For  self-contentedness, we reproduce an example similar to that of \citet{wehenkel2025addressing} that shows 
that a simulator may be well-specified according to the standard definition of misspecification
but still provide biased estimates of the target parameter when applied to real data.}

{
Consider a process that produces observations $y_1, \ldots y_N$ of some quantity  $\theta \in \mathbb{R}$ with a physical meaning of interest. The observations are realizations of a true DGP, $y \sim p_{\theta^*}={\cal N}(\theta^*, \sigma^2)$ where $\theta^*$ is the true value. Assume that instead of the true DGP, we have access to a simulator which produces for a given $\theta$, simulations that are realizations of $x \sim p_\theta = {\cal N}(\theta+\lambda, \sigma^2)$ where $\lambda >0$ is a fixed offset.
According to the standard definition, we are in a well-specified setting as we are in an ${\cal M}$-closed problem since there exists a value of $\theta= \theta^* -\lambda$ such that $p_\theta = p_{\theta^*}$. However, the posterior estimates with such a simulator are biased with respect to $\theta^*$. 
To see that, let us assume a prior ${\cal N}(\theta_0, \sigma^2_0)$ on $\theta$. The specific choice of prior is not important as its impact vanishes as the number of observations $N$ tends to infinity but the above choice is convenient to use Bayesian conjugacy properties. Indeed, it follows that $p(\theta | y_1, \ldots, y_N)$ is Gaussian with,
$$p(\theta | y_1, \ldots, y_N) = {\cal N}(\mu_N, \sigma^2_N)$$
where $\sigma^2_N = \left(\frac{1}{\sigma^2_0} + \frac{N}{\sigma^2}\right)^{-1}$ and $\mu_N = \sigma^2_N\left(\frac{\theta_0}{\sigma^2_0} + \frac{1}{\sigma^2} \sum_{i=1}^N (y_i -\lambda)\right)$.}

{Therefore,  when $N$ tends to infinity, 
$\mathbb{E}[\theta |y_1, \ldots, y_N] \rightarrow \mathbb{E}[y]-\lambda= \theta^*-\lambda$, showing  a bias that does not vanish. }

\section{Experimental Setup}\label{app:exp_setup}

We provide additional details on the training setup used across all experiments. 
For every task, we allocate a simulation budget of $N_{\text{sim}} = 5 \times 10^{4}$ samples and evaluate four calibration set sizes, $\ncal \in \{10,50,200,1000\}$, which we denote by $N_1,N_2,N_3,N_4$.

During training, $20\%$ of each calibration set is reserved for validation and the remaining $80\%$ is used for training. We did not optimize over random seeds. All models were trained on a Nvidia RTX3060 GPU under 3 hours. 

\textbf{Data preprocessing.} For tasks \TaskB, \TaskC, and \TaskD\ (uniform priors), we apply a logit transformation to map prior samples into $\thetaspace$. All datasets are standardized (z-scored) prior to training.

\textbf{Calibration sets.} For each $\ncal$, we generate 5 calibration sets by subsampling from a larger pool of calibration data. To reduce variance across runs, the sets are constructed in a nested fashion: denoting by $\mathcal{D}_{N_r}^i$ the $i$-th calibration set of size $N_r$, we enforce $\mathcal{D}_{N_r}^i \subset \mathcal{D}_{N_s}^i$ whenever $r < s$.

\textbf{Neural Posterior Estimation (NPE).} {For a fair comparison, the  procedure implemented in \ROPE\ is used for all compared methods. We rely on the UMNN flow implementation available at \href{https://github.com/AWehenkel/UMNN}{https://github.com/AWehenkel/UMNN}.} NPE is implemented with two components: a neural statistic estimator (NSE), $h_{\omega}(\mathbf{x})$, that encodes data into a low-dimensional representation, and a normalizing flow (NF) that maps a base distribution to the posterior. For task \TaskA, we use a standard neural spline flow~\citep{nsf} and omit the NSE. For tasks \TaskB, \TaskC\ and \TaskD, we reuse the architectures and hyperparameters from \citet{wehenkel2025addressing}.

\textbf{Flow Matching.} We use the architecture of \citet{dax_flow_2023} as a backbone. For the $\rvtheta$-space flow $\uth$, conditioning on $\mathbf{x}$ is implemented through a task-specific embedding network. For the data-space flow $\ux$, we employ the same architecture but with a separate embedding head for $(\mathbf{x}_t,t)$, equipped with positional encoding.

\textbf{Evaluation details.} 
The Wasserstein distance is computed using the \href{https://pypi.org/project/POT/}{\texttt{POT}} package with default settings. 
For the C2ST, we train a classifier based on an MLP backbone, augmented with an embedding network for $\mathbf{y}$; the embedding architecture matches the one used for the normalizing flow. 
We apply 3-fold cross-validation and report the average validation accuracy across folds. 
By default, we balance the two classes - $C=0$ (true samples) and $C=1$ (generated samples)—but note that stratified K-fold can also be used to handle class imbalance.

\subsection{Gaussian task}

The Gaussian task is defined as
\begin{equation}
    p_{\bm{\Theta}} = \mathcal{N}(\mu_\theta,\Sigma_\theta), 
    \quad p_{\bm{X}|\bm{\Theta}} = \mathcal{N}(A\bm{\theta} + b, \Sigma_x), 
    \quad p_{\bm{Y}|\bm{\Theta}} = \mathcal{N}(C\bm{\theta} + d, \Sigma_y),
\end{equation}
where
\[
\mu_\theta \in \R^3, \;\Sigma_\theta \in \R^{3\times 3}, \;
A \in \R^{10\times3}, \; b \in \R^{10}, \;\Sigma_x \in \R^{10\times10}, \;
C \in \R^{10\times3}, \; d \in \R^{10}, \;\Sigma_y \in \R^{10\times10}.
\]
All parameters above are drawn randomly at the start of the experiment. {The exact values can be found in the code by using the same seed we used for the experiment, which is hard-coded in the main script.}

\subsection{{Pendulum task}}

We follow the setup of \citet[Appendix I.2]{wehenkel2025addressing}. 
We sample $N=200$ timesteps $t_i \sim \mathcal{U}[0,10]$ and define the simulator as
\begin{equation}
S:
\begin{array}{rcl}
\bm{\theta}, \epsilon = (\eta_1, \dots, \eta_N, \varphi) & \longmapsto & [x_1,\dots,x_N]^T \\[0.5em]
\text{with } \;
x_i &=& A \cos(\omega_0 t_i + \varphi) + \eta_i,

\end{array}
\end{equation}
where $\varphi \sim \mathcal{U}[0,2\pi], \; \eta_i \sim \mathcal{N}(0,\sigma^2)$ {with $\sigma= 0.1$} and $\bm{\theta} = [A,\omega_0]$ are the parameters of interest that we try to infer.
The high-fidelity data-generating process (DGP) includes damping:
\begin{equation}
\text{DGP}:
\begin{array}{rcl}
\bm{\theta}, \epsilon = (\eta_1,\dots,\eta_N, \varphi, \alpha) & \longmapsto & [y_1,\dots,y_N]^T \\[0.5em]
\text{with } \;
y_i &=& e^{-\alpha t_i}\, A \cos(\omega_0 t_i + \varphi) + \eta_i,

\end{array}
\end{equation}
where $\alpha \sim \mathcal{U}[0,1], \; \; \varphi \sim \mathcal{U}[0,2\pi], \;\; \eta_i \sim \mathcal{N}(0,\sigma^2)$.  
The prior is uniform: $p_{\bm{\Theta}} = \mathcal{U}[0,3] \times \mathcal{U}[0.5,10]$.  

% For flow matching, we employ a residual MLP with GLU activations as in \citet{dax_flow_2023}, combined with the embedding network of \citet{wehenkel2025addressing} to obtain latent representations of $\mathbf{x}$ and $\mathbf{y}$.

\subsection{Wind Tunnel task}

We use the \texttt{load\_out\_0.5\_osr\_downwind\_4} experiment from the \texttt{wt\_intake\_impulse\_v1} dataset~\citep{Gamella2025}. The data consist of 50-step time series measuring air pressure in the chamber after an impulse applied to the input fan. A hatch on the side controls an additional opening, which can be controlled with precision. The inference task is to predict its position $H \in [0,45]$. We adopt model A2C3 from the \href{https://pypi.org/project/causalchamber/}{\texttt{causalchamber}} package as the simulator.

\subsection{Light Tunnel task}

We use the light tunnel experiment \texttt{uniform\_ap\_1.8\_iso\_500.0\_ss\_0.005} from the \texttt{lt\_camera\_v1} dataset \citep{Gamella2025}. A camera at the rear-end of an elongated chamber captures a light source emitted from the other end passing through two linear polarizers. We refer the reader to \citep{Gamella2025} for more details about the mechanistic model of the tunnel. The inference task consists in inferring the color of the light source ($(R,G,B) \in [0,255]^3)$ as well as the Malus law~\citep{Collett2005FieldGuide} coefficient $\alpha \in [0,1]$. The prior over these variables is uniform. The coefficient is a function of the polarizer angle $\alpha = \cos^2(\phi_1 - \phi_2)$, which are given in the dataset. The misspecification is introduced by omitting some physical aspects, which are detailed in \citep[Appendix D.IV.2.2]{Gamella2025}.

% For both tasks C (Wind Tunnel) and D (Light Tunnel), we used the same convolution architecture as in \citep{wehenkel2025addressing} to obtain latent representation for $\rvx$,$\rvy$. We also added positional encoding for the vector $(t, \rvx_t)$ when training $\ux$.

\section{Posterior Calibration}\label{app:sbc}

Simulation-Based Calibration (SBC) \citep{talts2020validating} is a principled method for 
assessing the internal consistency of an approximation $\hat p(\bm{\theta} \mid \bm{y})$ to the true posterior. 
It exploits the self-consistency property of Bayesian inference: if the approximate posterior 
is exact, then parameters  drawn from the prior should be uniformly 
distributed in rank among samples from the approximate posterior.
{Note however, that as many easy-to-deploy coverage procedures for SBI,  SBC does not provide sufficient conditions for posterior correctness. In particular, if the approximate posterior always returns the prior distribution instead of an approximation to the
posterior, then it passes the coverage diagnostic described below. Instead, SBC provides necessary conditions for posterior correctness, and thus enable detection of inaccuracies in the posterior estimate.
}
% see other methods in BDL book

The procedure is as follows.  For each pair $(\bm{\theta}_j, \bm{y}_j)$ from a test set $\mathcal{D}_{N} = \big\lbrace (\rvtheta_j,\rvy_j)\big\rbrace_{ 1 \leq j \leq N} $, we draw $L$ samples from the approximate 
posterior,
\begin{equation}
    \{\tilde{\bm{\theta}}^{(l)}_j\}_{1\leq l\leq L} \sim \hat{p}(\bm{\theta} \mid \bm{y}_j).
\end{equation}
Since $\bm{\theta}$ is multivariate, calibration is assessed per coordinate. For each 
dimension $k \in \{1, \ldots, d\}$, we compute the marginal rank statistic
\begin{equation}
    r_{j,k} = \sum_{l=1}^{L} \mathbf{1}\!\left[\tilde{\theta}^{(l)}_{j,k} < \theta_{j,k}
    \right] \in \{0, 1, \ldots, L\},
\end{equation}
where $\theta_{j,k}$ denotes the $k$-th component of $\bm{\theta}_j$. Under a well-calibrated 
posterior, the ranks $\{r_{j,k}\}_{1\leq j\leq N}$ should be uniformly distributed over 
$\{0, \ldots, L\}$ for every coordinate $k$. Systematic deviations from uniformity reveal 
specific failure modes: a U-shaped rank distribution indicates overdispersion (the approximate 
posterior is too wide), while a hump-shaped distribution indicates underdispersion (too narrow). 
A shift toward low or high ranks signals a global bias in the posterior mean.

In our experiments, we evaluate SBC using $N=1000$ calibration pairs and $L=500$ posterior samples per 
pair. We reported the results for both the \TaskA~ and \TaskC~ in Figures \ref{fig:sbc_gaussian} and \ref{fig:sbc_wt} {that show the respective histograms summarizing the ranks with a number of bins set to 20 using the rule recommended  by \citet{talts2020validating}.}
Results are reported for a specific training seed, and are not averaged over multiple draws of the calibration set, unlike the metrics displayed in Section \ref{sec:results}. 

% \begin{figure}[htbp]
%     \centering
%     \begin{subfigure}[b]{0.48\textwidth}
%         \includegraphics[width=\textwidth]{figures/calibration/gaussian/coverage_seed43_ncal10.pdf}
%         \caption{$n_\text{cal} = 10$}
%     \end{subfigure}
%     \hfill
%     \begin{subfigure}[b]{0.48\textwidth}
%         \includegraphics[width=\textwidth]{figures/calibration/gaussian/coverage_seed43_ncal50.pdf}
%         \caption{$n_\text{cal} = 50$}
%     \end{subfigure}

%     \vspace{1em}

%     \begin{subfigure}[b]{0.48\textwidth}
%         \includegraphics[width=\textwidth]{figures/calibration/gaussian/coverage_seed43_ncal200.pdf}
%         \caption{$n_\text{cal} = 200$}
%     \end{subfigure}
%     \hfill
%     \begin{subfigure}[b]{0.48\textwidth}
%         \includegraphics[width=\textwidth]{figures/calibration/gaussian/coverage_seed43_ncal1000.pdf}
%         \caption{$n_\text{cal} = 1000$}
%     \end{subfigure}

%     \caption{Coverage for varying calibration set sizes (seed 43).}
%     \label{fig:gaussian_coverage_seed43}
% \end{figure}

\begin{figure}[htbp]
    \centering
    \begin{subfigure}[b]{0.48\textwidth}
        \includegraphics[width=\textwidth]{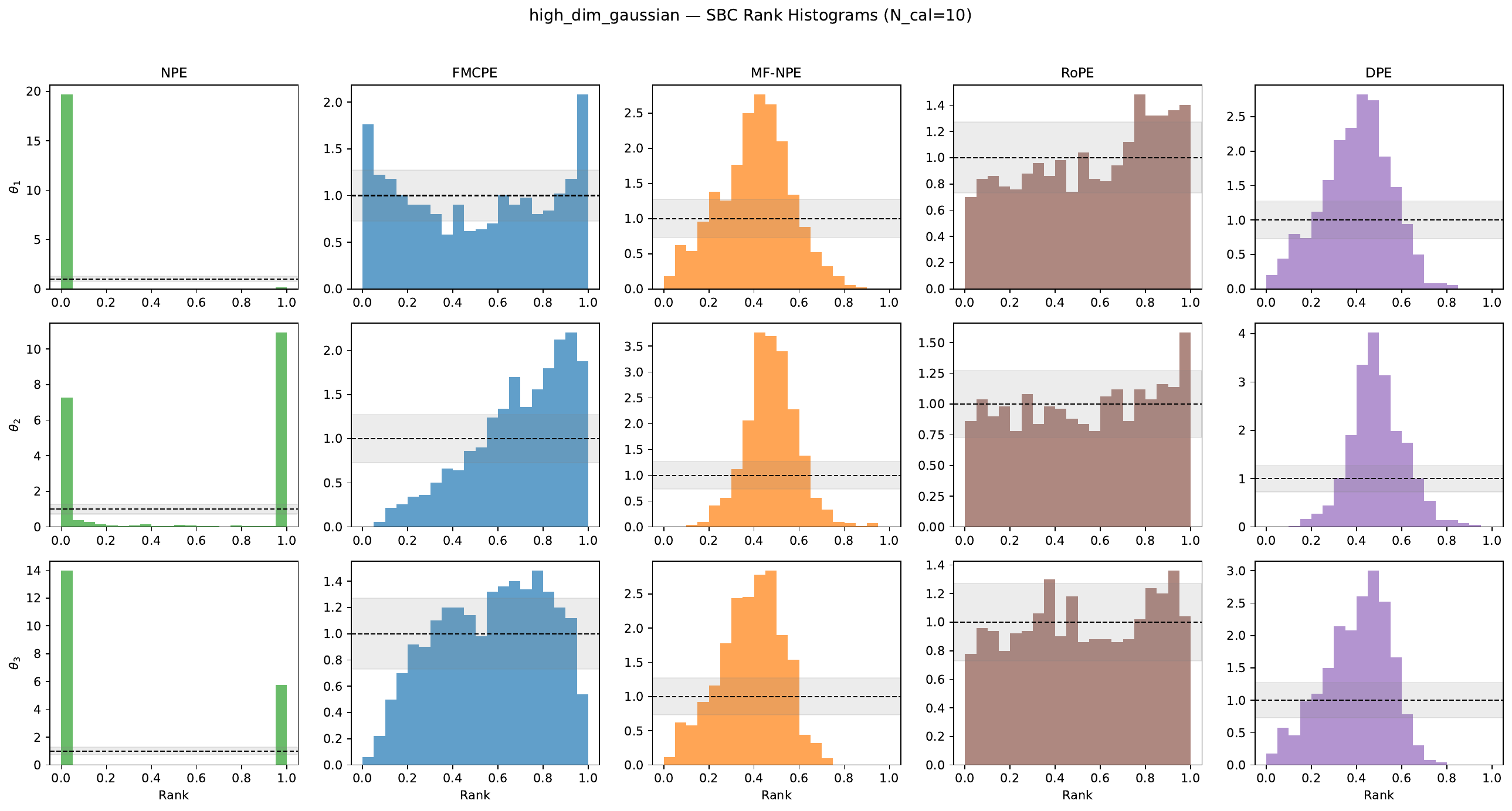}
        \caption{$\ncal=10$}
        \label{fig:sbc_gaussian_10}
    \end{subfigure}
    \hfill
    \begin{subfigure}[b]{0.48\textwidth}
        \includegraphics[width=\textwidth]{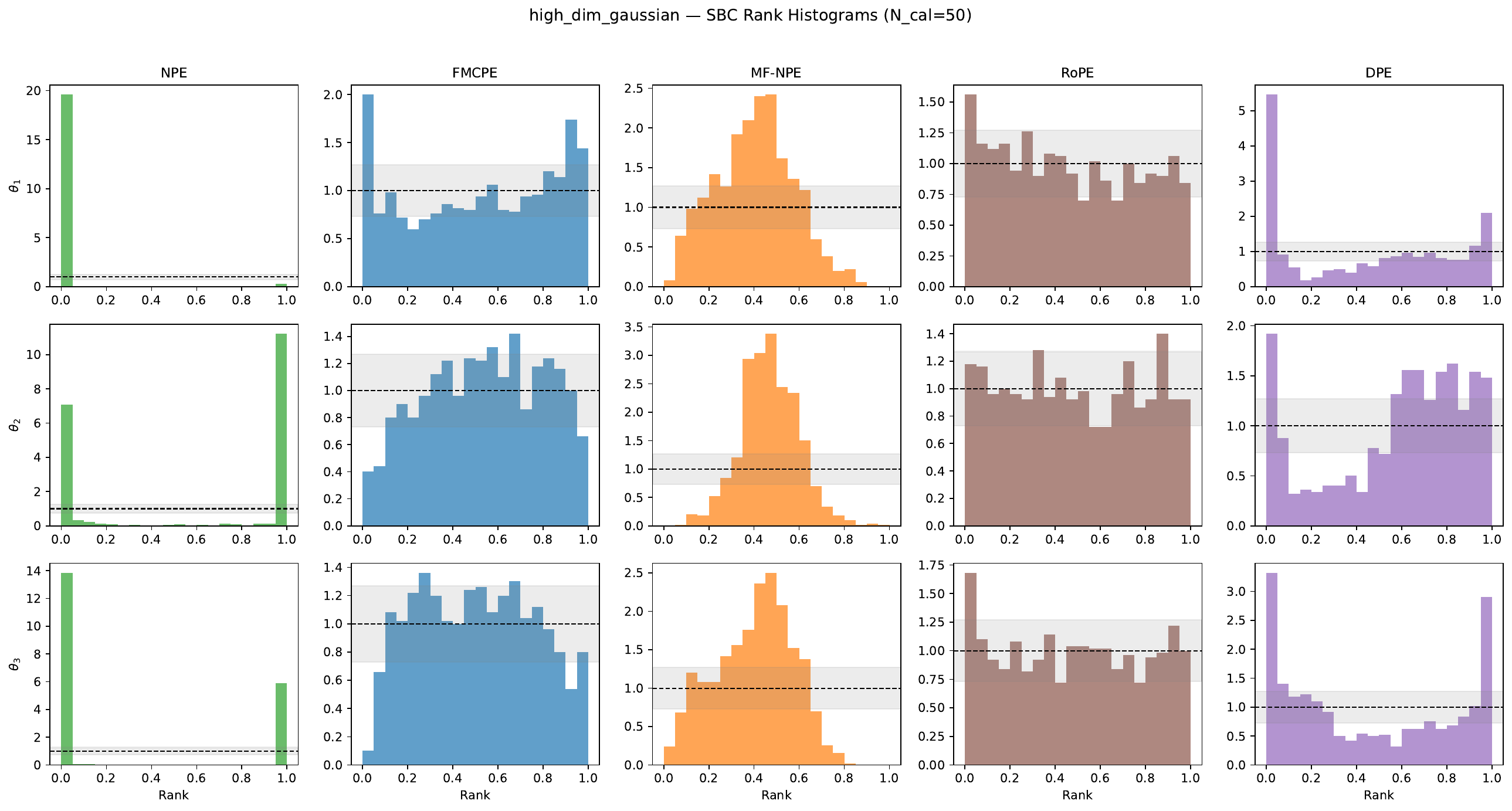}
        \caption{$\ncal=50$}
        \label{fig:sbc_gaussian_50}
    \end{subfigure}

    \vspace{1em}

    \begin{subfigure}[b]{0.48\textwidth}
        \includegraphics[width=\textwidth]{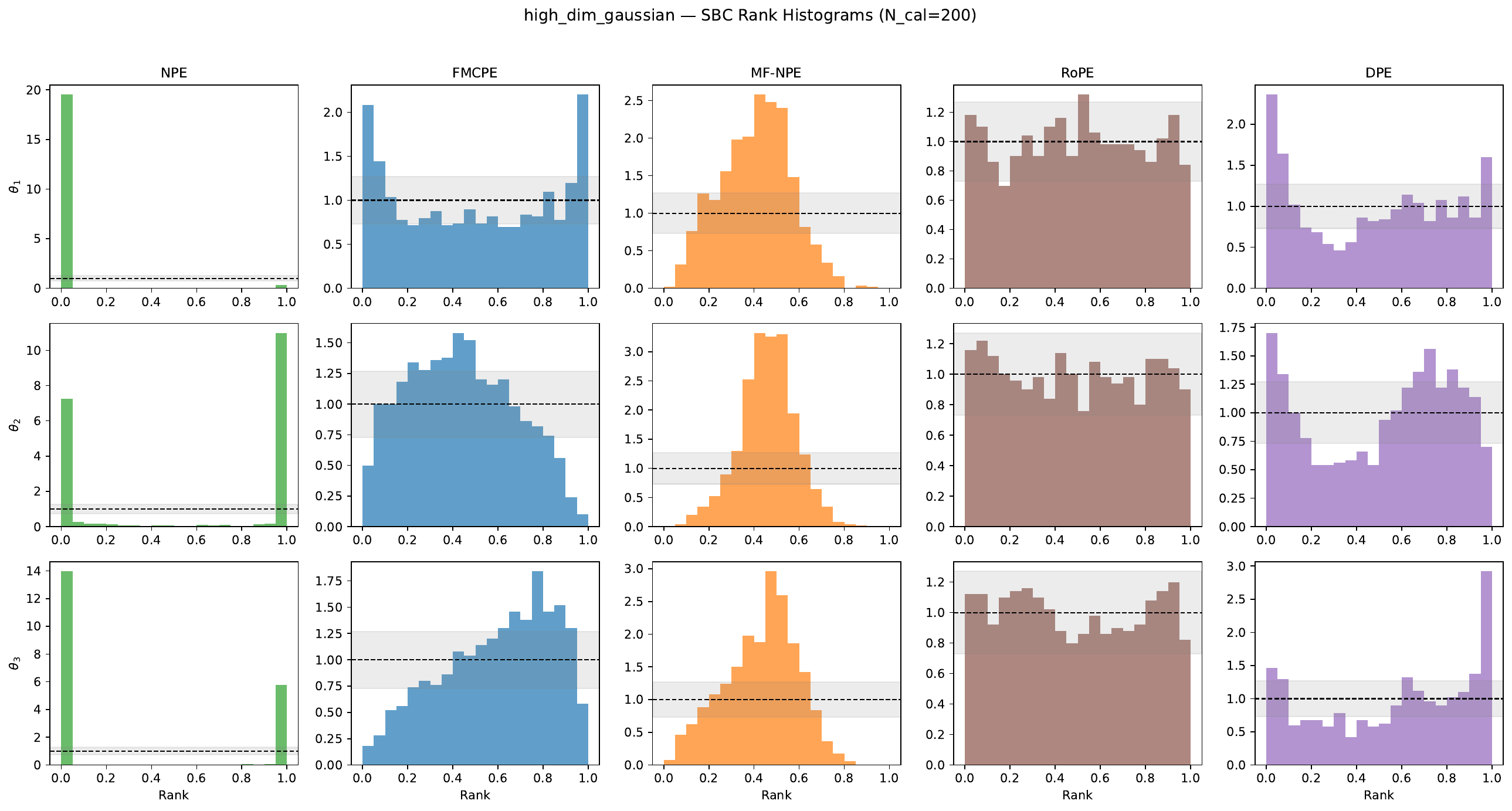}
        \caption{$\ncal=200 $}
        \label{fig:sbc_gaussian_200}
    \end{subfigure}
    \hfill
    \begin{subfigure}[b]{0.48\textwidth}
        \includegraphics[width=\textwidth]{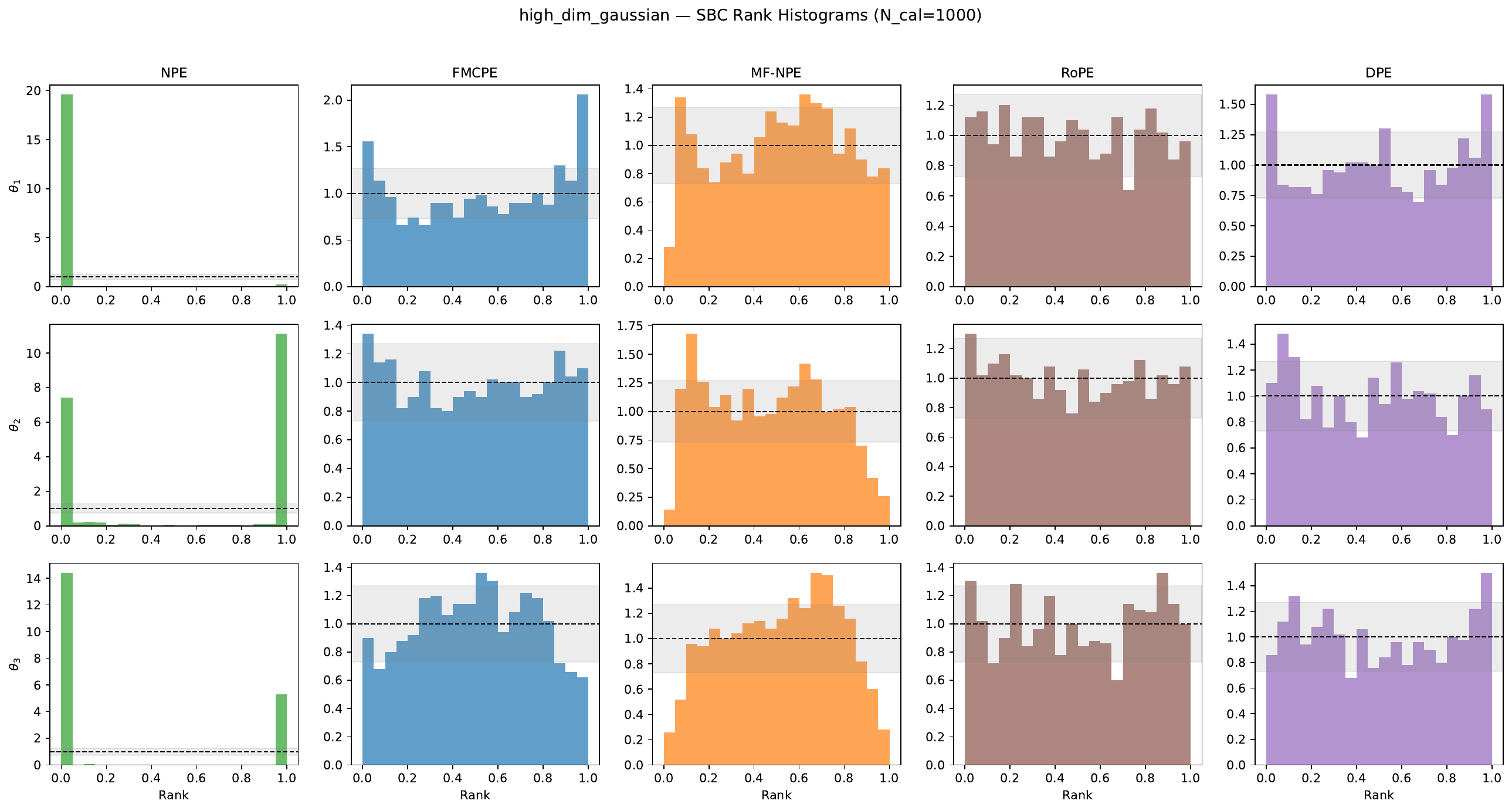}
        \caption{$\ncal=1000$}
        \label{fig:sbc_gaussian_1000}
    \end{subfigure}

    \caption{Simulation Based Calibration for the \TaskA~ task. {The four plots represent increasing $N_\text{cal}$ from 10 to 1000. In each plot, the rows correspond to the parameter dimensions $k=1,2,3$. The histograms then show the distributions of the $\{r_{j,k}\}_{1\leq j \leq N}$ for each method in columns, namely  NPE, \ourmethod, \MFNPE, \ROPE\ and DPE.} A perfectly calibrated method should yield a uniform distribution, represented by the dashed black line. The gray band represents the 95\% confidence interval of the bins distribution under the uniform assumption. }
    \label{fig:sbc_gaussian}
\end{figure}

% \begin{figure}[htbp]
%     \centering
%     \begin{subfigure}[b]{0.48\textwidth}
%         \includegraphics[width=\textwidth]{figures/calibration/wind_tunnel/coverage_seed43_ncal10.pdf}
%         \caption{$n_\text{cal} = 10$}
%     \end{subfigure}
%     \hfill
%     \begin{subfigure}[b]{0.48\textwidth}
%         \includegraphics[width=\textwidth]{figures/calibration/wind_tunnel/coverage_seed43_ncal50.pdf}
%         \caption{$n_\text{cal} = 50$}
%     \end{subfigure}

%     \vspace{1em}

%     \begin{subfigure}[b]{0.48\textwidth}
%         \includegraphics[width=\textwidth]{figures/calibration/wind_tunnel/coverage_seed43_ncal200.pdf}
%         \caption{$n_\text{cal} = 200$}
%     \end{subfigure}
%     \hfill
%     \begin{subfigure}[b]{0.48\textwidth}
%         \includegraphics[width=\textwidth]{figures/calibration/wind_tunnel/coverage_seed43_ncal1000.pdf}
%         \caption{$n_\text{cal} = 1000$}
%     \end{subfigure}

%     \caption{Coverage for varying calibration set sizes (seed 43).}
%     \label{fig:wt_coverage_seed43}
% \end{figure}

\begin{figure}[htbp]
    \centering
    \begin{subfigure}[b]{0.48\textwidth}
        \includegraphics[width=\textwidth]{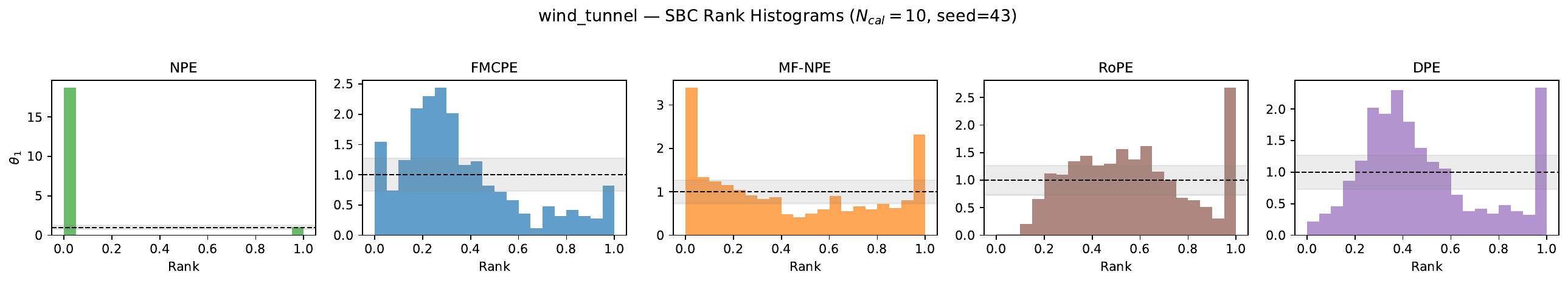}
        \caption{$\ncal=10$}
        \label{fig:sbc_wt_10}
    \end{subfigure}
    \hfill
    \begin{subfigure}[b]{0.48\textwidth}
        \includegraphics[width=\textwidth]{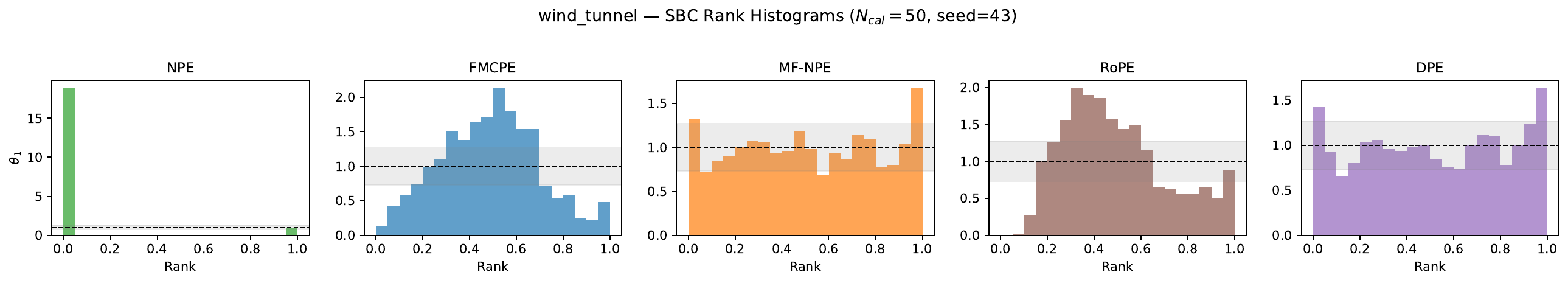}
        \caption{$\ncal=50$}
        \label{fig:sbc_wt_50}
    \end{subfigure}

    \vspace{1em}

    \begin{subfigure}[b]{0.48\textwidth}
        \includegraphics[width=\textwidth]{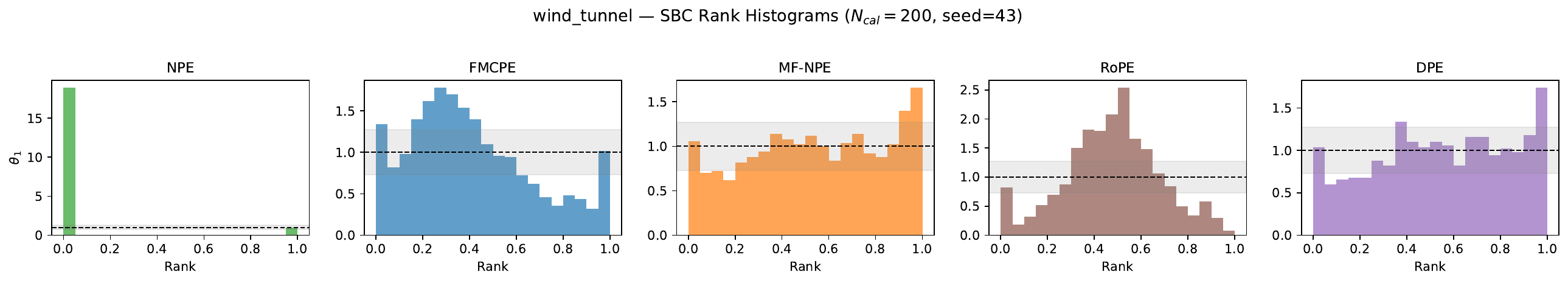}
        \caption{$\ncal=200$}
        \label{fig:sbc_wt_200}
    \end{subfigure}
    \hfill
    \begin{subfigure}[b]{0.48\textwidth}
        \includegraphics[width=\textwidth]{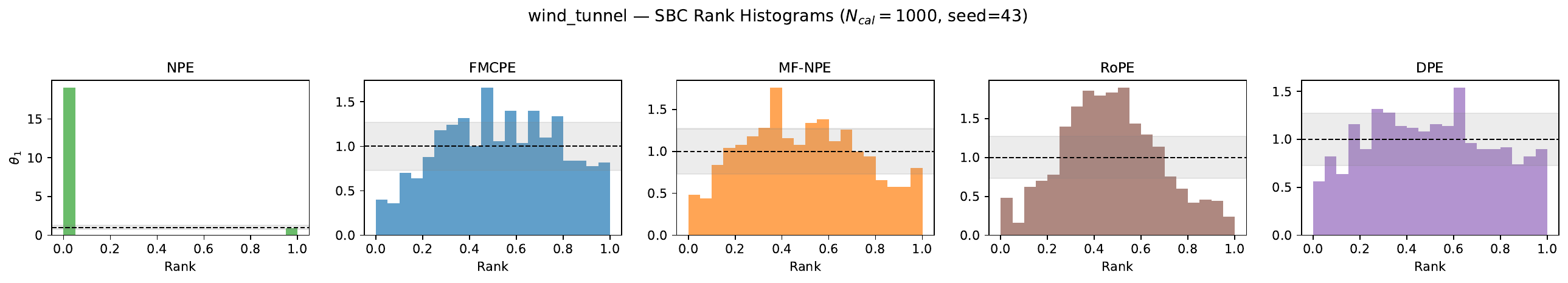}
        \caption{$\ncal=1000$}
        \label{fig:sbc_wt_1000}
    \end{subfigure}

    \caption{Simulation Based Calibration for the \TaskC~ task. {The four plots represent increasing $N_\text{cal}$ from 10 to 1000. In each plot, the histograms then show the distributions of the $\{r_{j}\}_{1\leq j \leq N}$ for the single parameter and  each method in columns, namely  NPE, \ourmethod, \MFNPE, \ROPE\ and DPE.} A perfectly calibrated method should yield a uniform distribution, represented by the dashed black line. The gray band represents the 95\% confidence interval of the bins distribution under the uniform assumption. }
    \label{fig:sbc_wt}
\end{figure}

For the \TaskA\ task, all methods tend toward a uniform rank distribution at $\ncal = 1000$, 
except for the NPE baselines, which correspond to a NPE posterior trained solely on simulation 
data and evaluated on held-out calibration data. At lower values of $\ncal$, our method still 
yields satisfactory calibration compared to DPE and \MFNPE, and is only outperformed by \ROPE. 
We attribute this behavior to the implicit regularization introduced by the parameters $\rho$ 
and $\gamma$ in the OT problem, which bias the resulting posterior toward the prior and 
therefore naturally promote calibrated rank distributions.

For the \TaskC\ task, all methods exhibit sub-optimal rank distributions even at $\ncal = 1000$. 
This may indicate either that the task is too challenging for the methods presented here to 
achieve satisfactory calibration, or that the SBC diagnostics were computed with insufficiently 
large values of $N$ and $L$. Since the observations are higher-dimensional than in the \TaskA\ 
task, we suspect that calibration could be better assessed given access to a larger number of 
test pairs $\{(\bm{\theta}_j, \bm{y}_j)\}_{1\leq j \leq N}$.

\section{Additional results}\label{app:plots_metrics}
\subsection{ Additional baselines and metrics}
 %In Figure~\ref{fig:all_metrics_all_methods} 
 Additional metrics (MSE and MMD) are displayed for all previously compared methods, as well as for the following two additional baselines from \citet{wehenkel2025addressing},
\begin{itemize}
    \item \textbf{Noisy Neural Posterior Estimation (NNPE)}: The amortized version of RNPE from \citet{ward_robust_2022}, where the authors added Spike and Slab noise to the simulations and then trained a NPE model on these noisy data. This allegedly makes the method more robust to model misspecification by providing more conservative posterior estimation.

    \item \textbf{Joint Neural Posterior Estimation (J-NPE):} A posterior estimator was trained on both simulated and calibration data jointly, $\mathcal{D}_{\text{train}} = \left\lbrace (\rvtheta_i,\rvx_i) \right\rbrace_{1\leq i \leq N_{\text{sim}}} \ \bigcup \ \left \lbrace (\rvtheta_j,\rvy_j)\right \rbrace_{1\leq j \leq N_{\text{cal}}}$.
\end{itemize}
As the ground truth posterior distribution is available for the Gaussian task,  posterior-level metrics are computed and reported in Figure~\ref{fig:gaussian_con_all_metrics}, namely,
\begin{itemize}
    \item \textbf{C2ST}: The regular C2ST as introduced in \citep{lopez-paz2017revisiting}.
    \item $\bm{W_2-\theta}$: The Wasserstein distance between parameter samples 
    $\{\tilde{\rvtheta}_i\}_{1 \leq i\leq M}$ 
    from an estimated posterior $\tilde{p}(\rvtheta | \rvy^*)$ and samples  $\{\rvtheta_i\}_{1 \leq i\leq M'}$  from the true posterior $p(\rvtheta | \rvy^*)$  for a given observation $\rvy^*$.
    \begin{align*}
\begin{aligned}
W_2 =\min_{\gamma \in \mathbb{R}_+^{M \times M'} }\left( \sum_{i,j}\gamma_{i,j}\|\rvtheta_i - \tilde{\rvtheta}_j\|^2_2\right)^\frac{1}{2}~\text{such that}~\gamma .\mathbf{1} = \tfrac{1}{M}~\text{and}~\gamma^T.\mathbf{1}= \tfrac{1}{M'} \; .
\end{aligned}
\end{align*}
\item \textbf{MMD}$\bm{-\theta}$: The Maximum Mean Discrepancy with a radial basis function (rbf) kernel between parameter samples 
    $\{\tilde{\rvtheta}_i\}_{1 \leq i\leq M}$ 
    from an estimated posterior $\tilde{p}(\rvtheta | \rvy^*)$ and samples  $\{\rvtheta_i\}_{1 \leq i\leq M'}$  from the true posterior $p(\rvtheta | \rvy^*)$  for a given observation $\rvy^*$.
\end{itemize}
For all other tasks, metrics are computed between joint distribution samples and reported in
Figure~\ref{fig:all_metrics_all_methods}.

Additionally, KDE plots for tasks \TaskC\ and \TaskD\ are shown, respectively, in Figures~\ref{fig:kde_plot_wind_tunnel} and \ref{fig:kde_plot_light_tunnel}. \ROPE\ was omitted in Figure~\ref{fig:kde_plot_light_tunnel} as it hindered clear visualization of the other posterior distributions due to a too flat posterior estimate.

{Another qualitative illustration of the impact of our proposed correction is given in Figures \ref{fig:before_after_gaussian} to \ref{fig:before_after_lt}. The plots clearly show how the posteriors are shifted to a more appropriate location after calibration.}

% \begin{figure}[hbtp]
%     \centering
%     \includegraphics[width=\linewidth]{figures/mse_boxplots.pdf}
%     \caption{MSE with respect to an increasing calibration set size $N_{\text{cal}}\in \{10,50,200,1000\}$. Each boxplot shows the distribution of MSE values across five independent runs, each using a different randomly chosen calibration set.}
%     \label{fig:mse}
% \end{figure}

\begin{figure}[h]
    \centering
    \includegraphics[width=\linewidth]{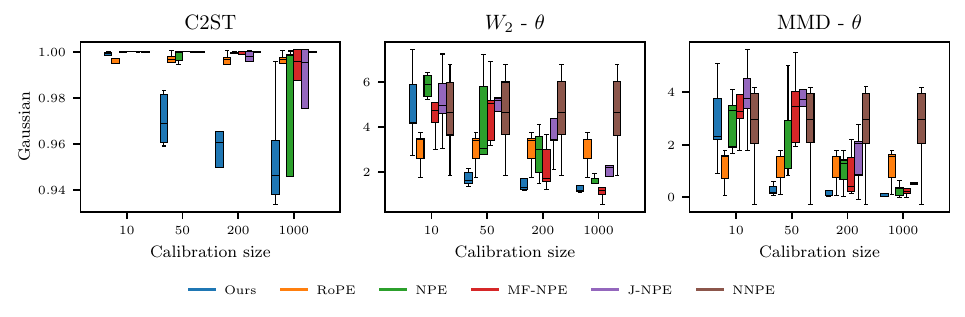}
    \caption{
    Parameter sample metrics ($\boldsymbol{\downarrow}$ is better).
    C2ST, Wasserstein distance and Maximum Mean Discrepancy (MMD) for the \TaskA\ task. For a given observation $\rvy^*$, we sampled $M =5000 $ parameters $\{\tilde{\rvtheta}_i\}_{1 \leq i\leq M}$ for each method. Each boxplot show the distribution of metric values over 3 different observations $\rvy^* \in {\cal D}_{\text{test}}$ chosen at random.}
    \label{fig:gaussian_con_all_metrics}
\end{figure}

\begin{figure}
    \centering
    \includegraphics[width=\linewidth]{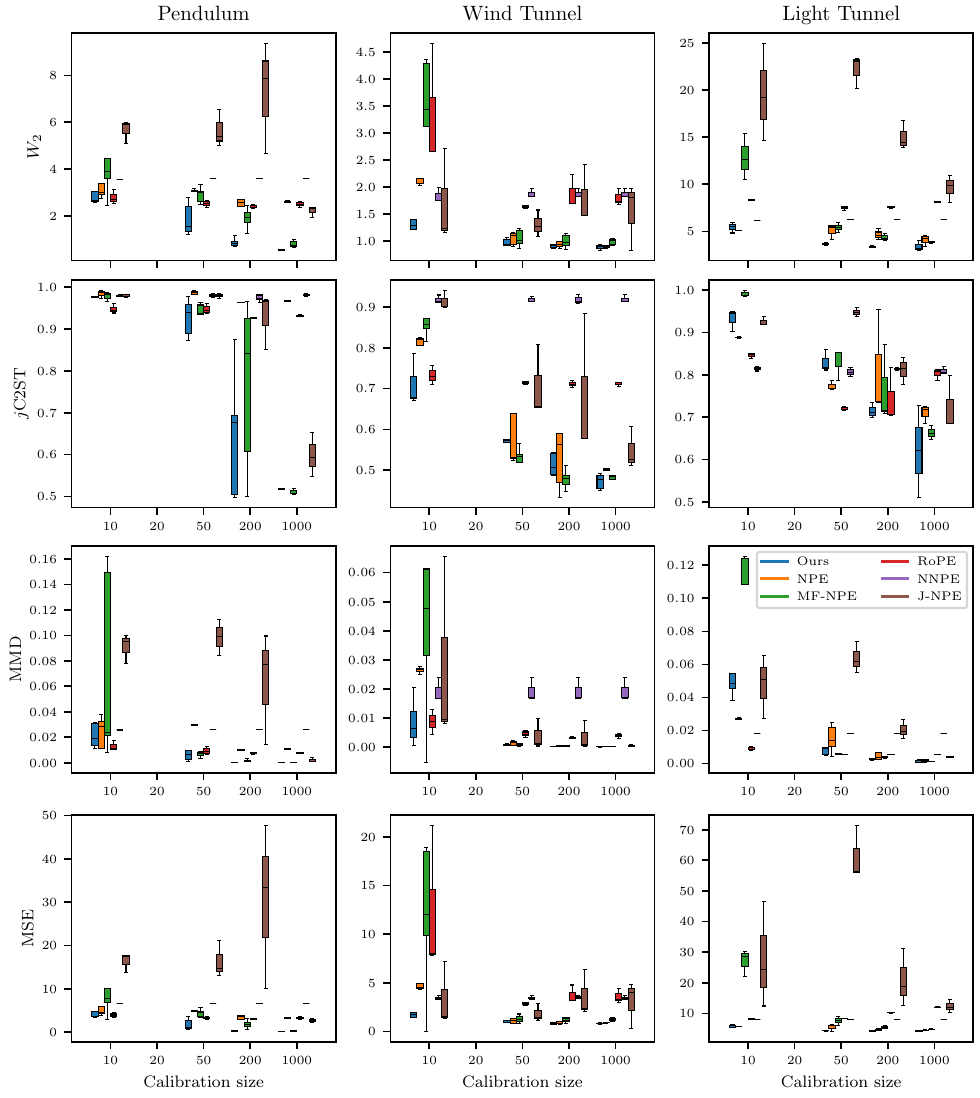}
    \caption{\footnotesize Joint sample metrics ($\boldsymbol{\downarrow}$ is better). Wasserstein distance (top row), $j$C2ST (second row) MMD (third row) and MSE (bottom row) with respect to an increasing calibration set size $N_{\text{cal}}\in \{10,50,200,1000\}$. Each boxplot shows the distribution of metric values across five independent runs, each using a different randomly chosen calibration set.}
    \label{fig:all_metrics_all_methods}
\end{figure}

\begin{figure}[btp]
    \centering

    \begin{subfigure}[b]{0.36\textwidth}
        \centering
        \includegraphics[width=\linewidth]{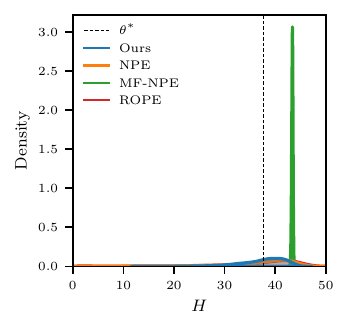}
        \caption{$N_{\text{cal}}=10$}
        \label{subfig:kde_plot_wind_tunnel_10}
    \end{subfigure}
    \hspace{-5pt}
    % Second subfigure
    \begin{subfigure}[b]{0.305\textwidth}
        \centering
        \includegraphics[width=\linewidth]{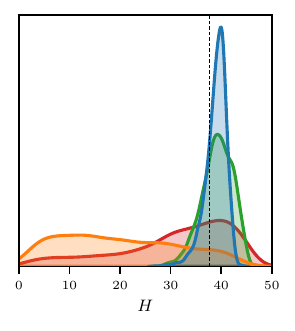}
        \caption{$N_{\text{cal}}=50$}
        \label{subfig:kde_plot_wind_tunnel_50}
    \end{subfigure}
    \hspace{-5pt}
    % Third subfigure
    \begin{subfigure}[b]{0.305\textwidth}
        \centering
        \includegraphics[width=\linewidth]{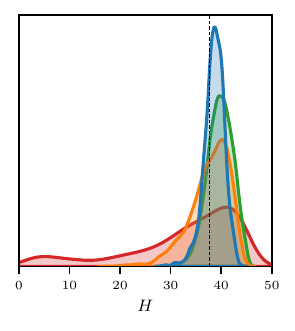}
        \caption{$N_{\text{cal}}=200$}
        \label{subfig:kde_plot_wind_tunnel_200}
    \end{subfigure}
    
    \caption{\label{fig:kde_plot_wind_tunnel}Kernel density estimates of the learned posteriors for task \TaskC. For a given  $\bm{y}^* \!\!\in \!{\cal D}_{\text{test}}$, we draw $\{\tilde{\bm{\theta}}_i\}_{1\leq i \leq 2000}$, for each method and 3 calibration sizes $N_{\text{cal}} \!\in\!\{ 10,50,200\}$. The dotted black line indicates the true parameter $\rvtheta^*$ that generated $\rvy^*$.}
\end{figure}

\begin{figure}[btp]
    \centering

    \begin{subfigure}[b]{0.365\textwidth}
        \centering
        \includegraphics[width=\linewidth]{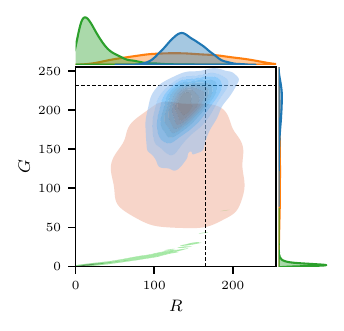}
        \caption{$N_{\text{cal}}=10$}
        \label{subfig:kde_plot_light_tunnel_10}
    \end{subfigure}
    \hspace{-5pt}
    % Second subfigure
    \begin{subfigure}[b]{0.305\textwidth}
        \centering
        \includegraphics[width=\linewidth]{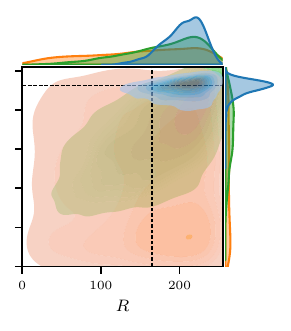}
        \caption{$N_{\text{cal}}=50$}
        \label{subfig:kde_plot_light_tunnel_50}
    \end{subfigure}
    \hspace{-5pt}
    % Third subfigure
    \begin{subfigure}[b]{0.305\textwidth}
        \centering
        \includegraphics[width=\linewidth]{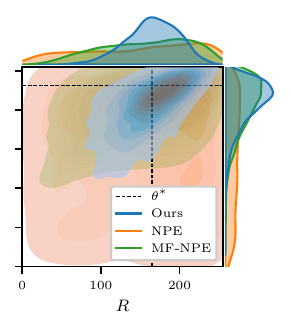}
        \caption{$N_{\text{cal}}=200$}
        \label{subfig:kde_plot_light_tunnel_200}
    \end{subfigure}
    
    \caption{\label{fig:kde_plot_light_tunnel}Kernel density estimates of joint and marginal samples for task \TaskD. We report the posterior densities for the first two coordinates of the parameter $[\theta_1,\theta_2] = (R,G)$. For a given  $\bm{y}^* \!\!\in {\cal D}_{\text{test}}$, we draw $\{\tilde{\bm{\theta}}_i\}_{1\leq i \leq 2000}$, for each method and 3 calibration sizes $N_{\text{cal}} \!\in\!\{ 10,50,200\}$. Dotted black lines indicate the true parameter $\rvtheta^*$ that generated $\rvy^*$.}
\end{figure}

% before and after Figures here
\begin{figure}
    \centering
    \includegraphics[width=0.5\linewidth]{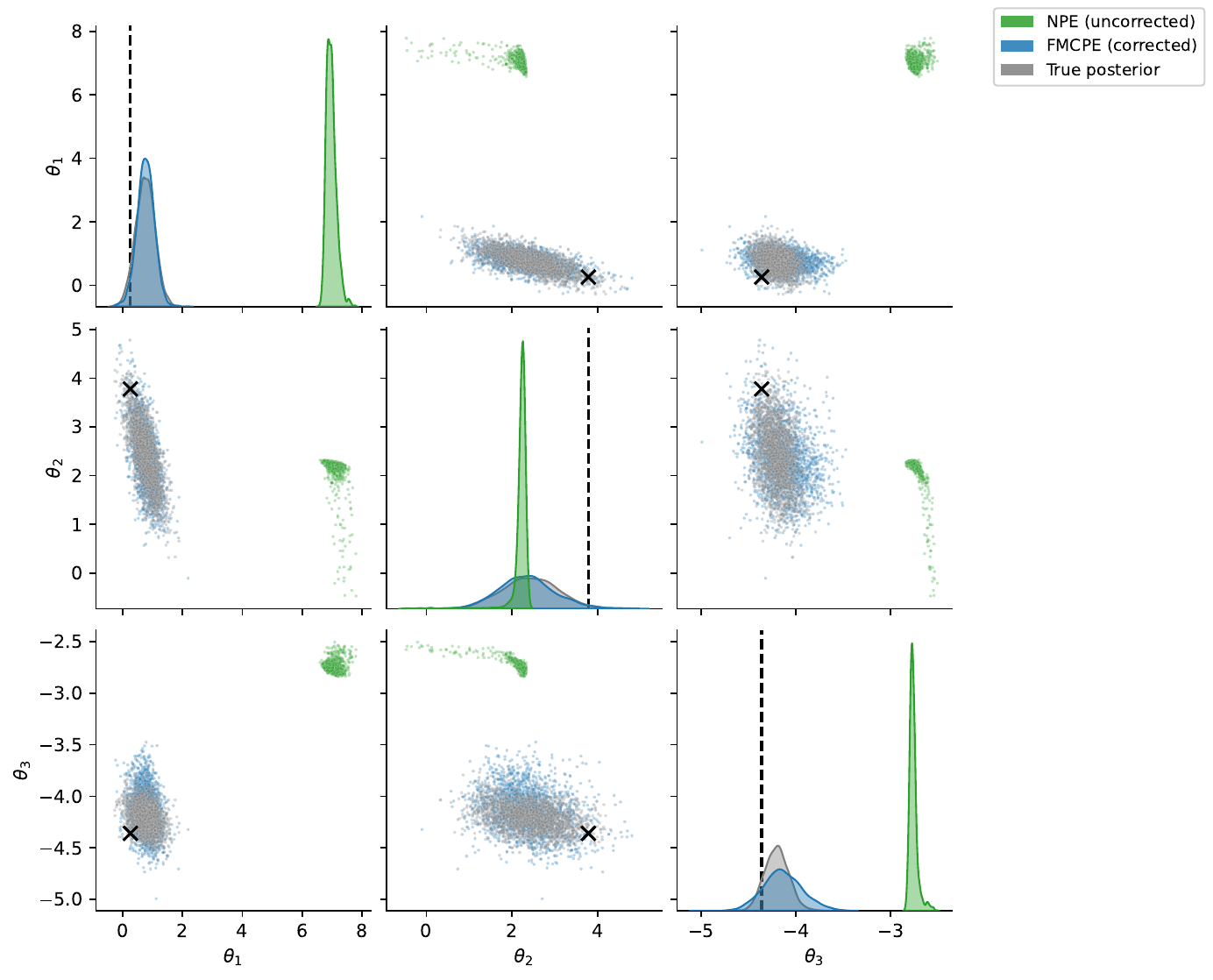}
    \caption{\TaskA~ task:  for a given observation, posterior distributions before (NPE in green) and after our \ourmethod\ correction (in blue) when $\ncal=1000$. True posterior in grey.}
    \label{fig:before_after_gaussian}
\end{figure}

\begin{figure}
    \centering
    \includegraphics[width=0.5\linewidth]{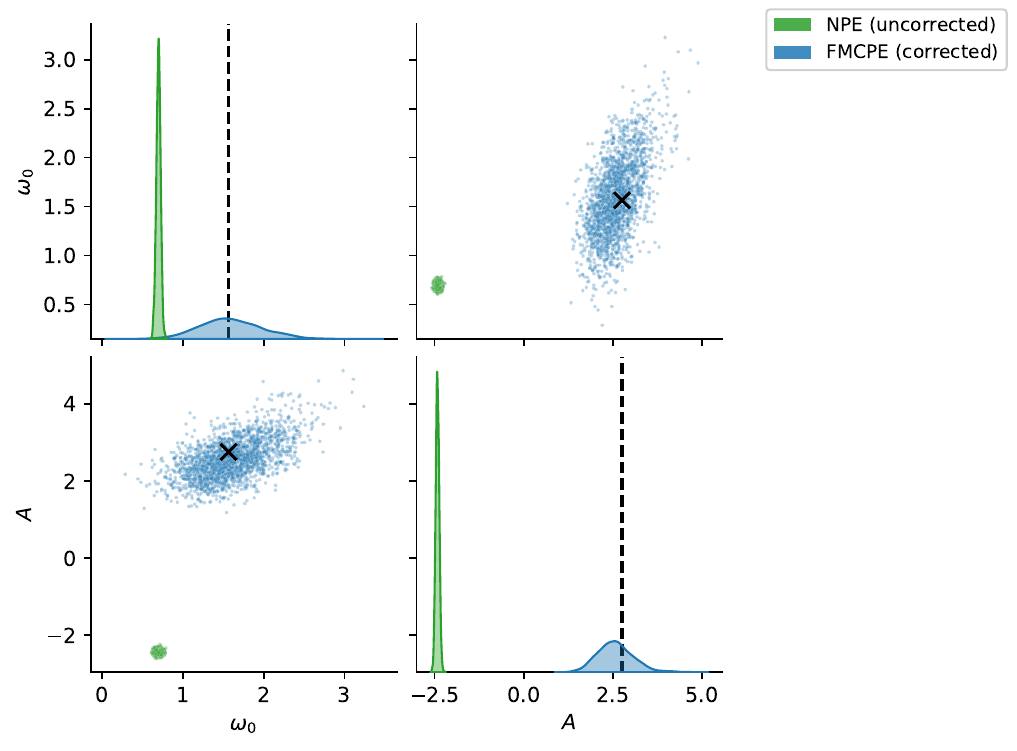}
    \caption{\TaskB\ task:
      for a given observation, posterior distributions before (NPE in green) and after our \ourmethod\ correction (in blue) when $\ncal=1000$.}
    \label{fig:before_after_pendulum}
\end{figure}

\begin{figure}
    \centering
    \includegraphics[width=0.5\linewidth]{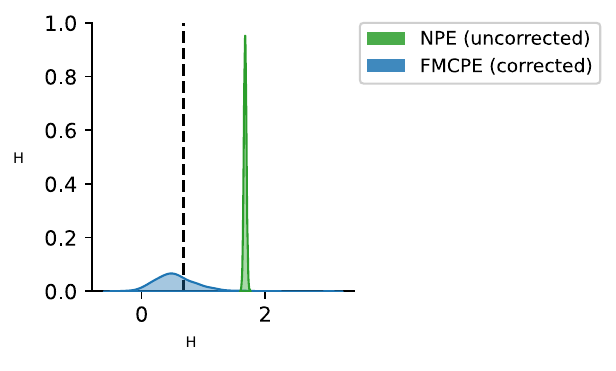} 
    \caption{\TaskC\ task: for a given observation, posterior distributions before (NPE in green) and after our \ourmethod\ correction (in blue) when $\ncal=1000$.
    }
    \label{fig:before_after_wt}
\end{figure}

\begin{figure}
    \centering
    \includegraphics[width=0.5\linewidth]{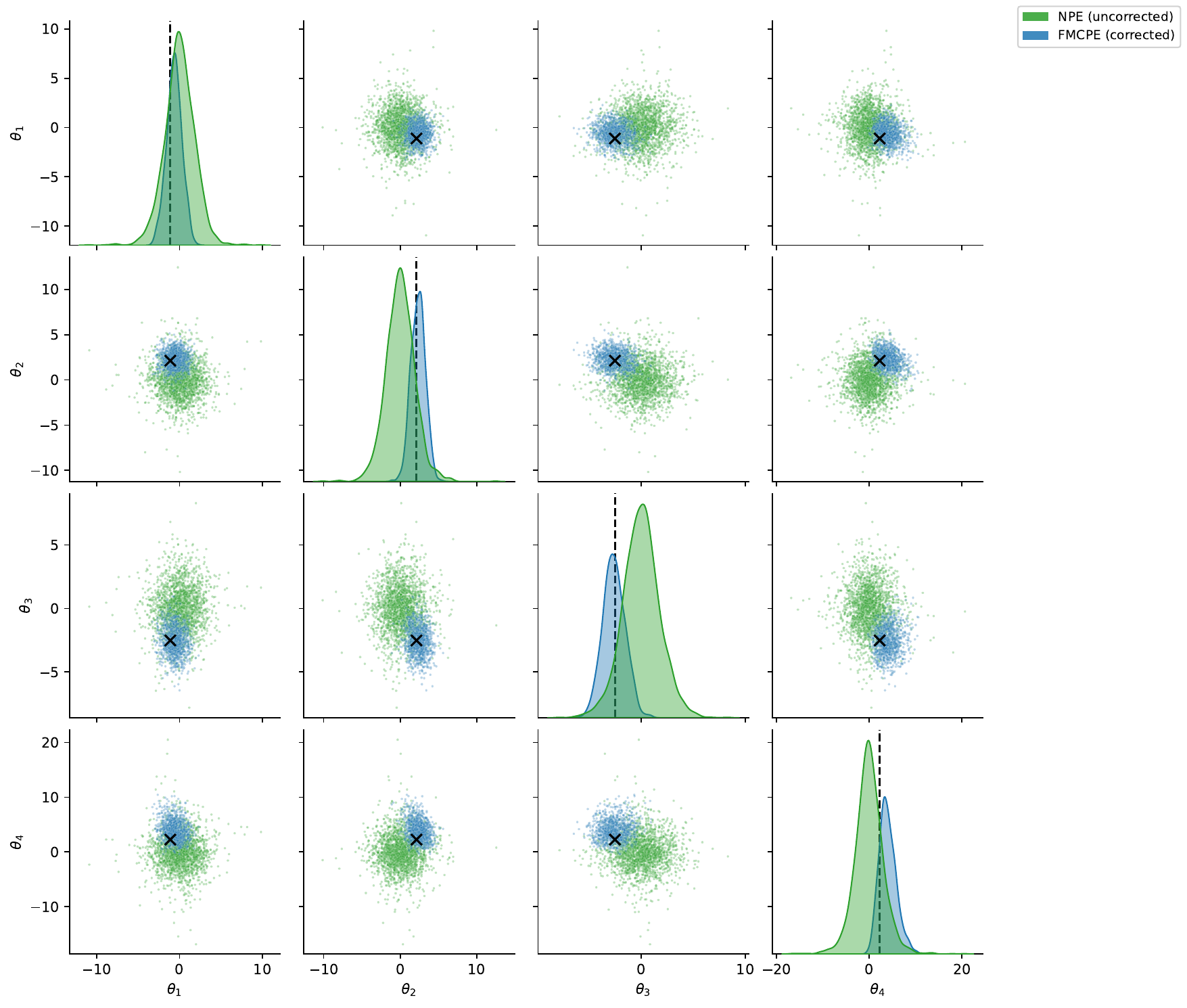}
    \caption{\TaskD\ task: for a given observation, posterior distributions before (NPE in green) and after our \ourmethod\ correction (in blue) when $\ncal=1000$. }
    \label{fig:before_after_lt}
\end{figure}

\subsection{Ablation Study: impact of joint learning}
\label{app:ablation}
To provide more insight on  the different \ourmethod\ components, 
 an ablation study is also conducted. Our joint \ourmethod\ training, referred to as $T_{\bm{\Theta}} + T_{\bm{X}}$ is first compared to a sequential one where the $T_{\bm{X}}$  map is  first learned to provide the source distribution for $T_{\bm{\Theta}}$ fitted in turns.
Figure \ref{fig:ablation}  shows that in all tested examples and metrics, the joint training provides better or equivalent performance, confirming the advantage of adjusting both maps in an interactive manner. 
To explore then the impact of each map separately, two options are considered. In the $T_{\bm{X}}$-only procedure,  NPE is used 
with  $\hat{p}_{\rvTheta | \rvX}(\rvtheta | \rvX= T_{\bm{X}}(\bm{y}) )$ where $T_{\bm{X}}$ is trained on calibration data,
%
%This results in a scheme similar in spirit to the RNPE procedure of \cite{ward_robust_2022} but using 
%MCMC for posterior corrections rather than a learned source.
%
while $T_{\bm{\Theta}}$-only denotes a learned $T_{\bm{\Theta}}$ on calibration data from a source distribution $\pi(\bm{\theta} | \bm{y}) = \hat{p}_{\bm{\Theta}|\bm{X}}(\bm{\theta} | \bm{X} = \bm{y})$ ignoring calibration data.
The worse performance of $T_{\bm{X}}$-only, including when calibration data increases, suggests that calibration has more impact when learning $T_{\bm{\Theta}}$  than 
$T_{\bm{X}}$. 
In contrast, $T_{\bm{\Theta}}$-only and $T_{\bm{\Theta}} + T_{\bm{X}}-\text{Sequential}$, which differ mainly in the nature of the source, show similar performance. This suggests that a key feature of \ourmethod\ is the possibility  to  learn jointly  the source and  $T_{\bm{\Theta}}$ rather than the nature of the source  itself. 
Note that the \TaskA~example is not shown because  no significant differences were observed, which suggests that for simple models a sub-optimal use of calibration data may not be critical. 

\begin{figure}[t]
    \centering    
    \includegraphics[width=\linewidth]{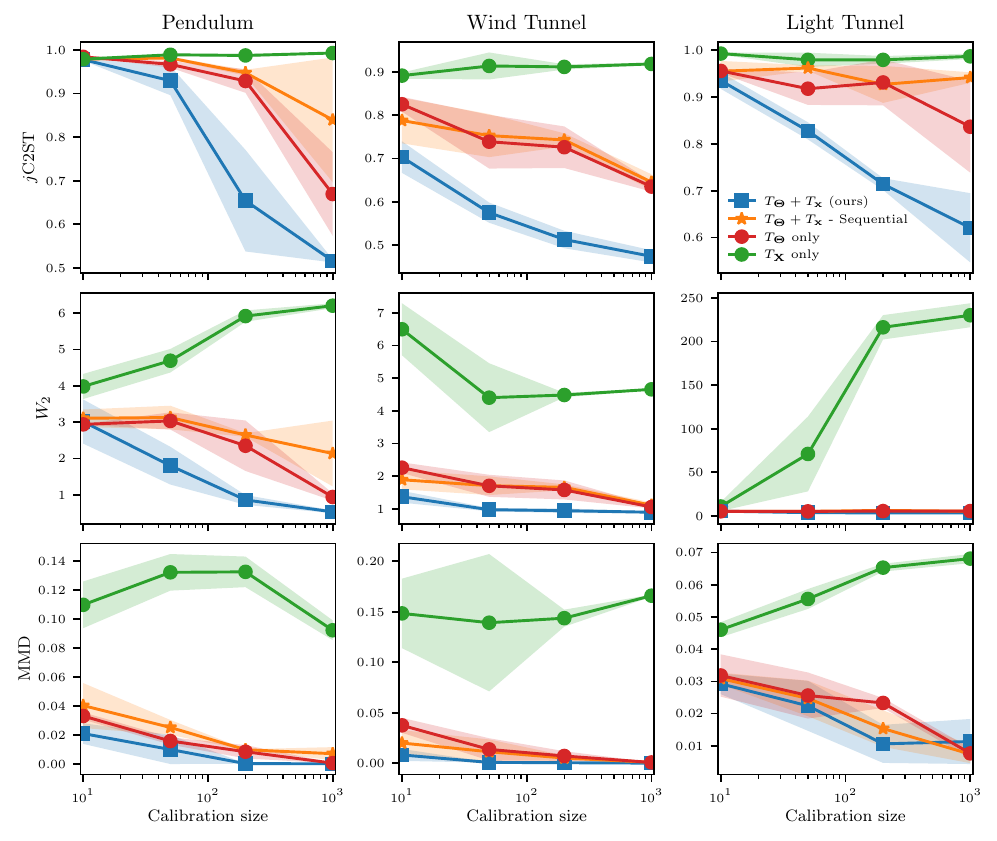}
    \caption{\label{fig:ablation}
    Ablation study investigating the impact of each \ourmethod\ component (the lower the better).
    Our joint $T_{\bm{\Theta}} + T_{\bm{X}}$ (blue) is compared to a sequential version (orange) learning $T_{\bm{X}}$ then $T_{\bm{\Theta}}$, and to two other options. In $T_{\bm{X}}$ only (green), resp. $T_{\bm{\Theta}}$ only  (red), calibration data is used only in learning $T_{\bm{X}}$, resp. $T_{\bm{\Theta}}$. 
    }
    \vspace{-1em}
\end{figure}

%\subsection{Before \& after calibration comparison}
%\label{app:before_after}

\subsection{{Impact of the conditional independence hypothesis}}
\label{app:hyp_test}
The conditional independence (CI) hypothesis (Equation~\ref{eq:hyp}) is not a requirement for our method to work in general.  We provide results on a toy Gaussian example where we control to which degree the CI holds with a scalar $\alpha$
\begin{align}
    &\bm{\theta} \sim \mathcal{U}([-3,3]^5), \quad \bm{\theta} \in \mathbb{R}^5 \\
    &\bm{x} \mid \bm{\theta} \sim \mathcal{N}(A\bm{\theta} , \sigma_x^2 I), \quad \bm{x} \in \mathbb{R}^{20},\ A \in \mathbb{R}^{20 \times 5} \\
    &\bm{y} \mid \bm{\theta}, \bm{x} \sim \mathcal{N}(C\bm{x} + \alpha D\bm{\theta} , \sigma_y^2 I), \quad C \in \mathbb{R}^{20 \times 20},\ D \in \mathbb{R}^{20 \times 5}
\end{align}
where $\sigma_x = 0.5$, $\sigma_y = 0.3$.
When $\alpha = 0$, $\bm{y}$ depends on $\theta$ only through $\bm{x}$ (CI holds); when $\alpha > 0$, the term $\alpha D\bm{\theta}$ introduces a direct $\bm{\theta} \to \bm{y}$ effect. We provide metrics values for \ourmethod\ using both $T_{\bm{\Theta}}$ and $T_{\bm{X}}$ and \ourmethod\ using only $T_{\bm{X}}$ in Table \ref{tab:ablation_flow_theta}. We can clearly see that the gap between both variants is small for $\alpha=0$, suggesting the approach using only $T_{\bm{X}}$ can be sufficient. The gap widens as $\alpha$ increases, confirming the positive impact of the $\bm{\theta}$-flow correction when conditional independence is violated.

\begin{table}[h!]
    \centering

    \begin{tabular}{clcccc}
        \toprule
        $\alpha$ & Method & jC2ST ($\downarrow$ 0.5) & $\Delta$\% & jMMD ($\downarrow$ 0) & $\Delta$\% \\
        \midrule
        \multirow{2}{*}{0.0}
            & \ourmethod\ (full)         & $.576 \pm .006$ &        & $.00040 \pm .00001$ &         \\
            & \ourmethod\ ($T_{\bm{X}}$ only)   & $.613 \pm .003$ & +6.4\% & $.00043 \pm .00001$ & +7.5\%  \\
        \midrule
        \multirow{2}{*}{0.5}
            & \ourmethod\ (full)         & $.630 \pm .015$ &        & $.00045 \pm .00000$ &         \\
            & \ourmethod\ ($T_{\bm{X}}$ only)   & $.673 \pm .004$ & +6.8\% & $.00051 \pm .00001$ & +13.3\% \\
        \midrule
        \multirow{2}{*}{1.0}
            & \ourmethod\ (full)         & $.691 \pm .016$ &         & $.00030 \pm .00003$ &         \\
            & \ourmethod\ ($T_{\bm{X}}$ only)   & $.777 \pm .003$ & +12.4\% & $.00048 \pm .00001$ & +60.0\% \\
        \midrule
        \multirow{2}{*}{2.0}
            & \ourmethod\ (full)         & $.768 \pm .041$ &         & $.00022 \pm .00003$ &          \\
            & \ourmethod\ ($T_{\bm{X}}$ only)   & $.864 \pm .004$ & +12.5\% & $.00045 \pm .00001$ & +104.5\% \\
        \midrule
        \multirow{2}{*}{5.0}
            & \ourmethod\ (full)         & $.545 \pm .051$ &         & $.00033 \pm .00017$ &         \\
            & \ourmethod\ ($T_{\bm{X}}$ only)   & $.709 \pm .168$ & +30.1\% & $.00052 \pm .00008$ & +57.6\% \\
        \bottomrule
    \end{tabular}
        \caption{Flow-$\bm{\theta}$ ablation, $N_{\text{cal}} = 1000$. Mean $\pm$ std over 3 seeds. The gap between \ourmethod\ (full) and \ourmethod\  ($T_{\bm{X}}$ only) is reported relatively to the value of \ourmethod\  (full) for each line.}
    \label{tab:ablation_flow_theta}
\end{table}

\subsection{Prior Misspecification}
\label{app:prior_misspec}
We conducted an additional experiment to check how our method behaved when the source of misspecification comes from the prior. We considered the following modifications of two of the previous tasks:

\begin{itemize}
    \item \TaskA\ task : We use  a modified prior, which is shifted and scaled compared to the true prior $p(\rvtheta)$. 
    $$ p_{\text{miss}}(\rvtheta) = \mathcal{N}(\mu + \delta, \alpha \Sigma), \quad  \mbox{while} \quad p(\rvtheta) = \mathcal{N}(\mu, \Sigma) $$
    where $\mu \in \mathbb{R}^3$ and $\Sigma \in \mathbb{R}^{3\times3}$ are the mean and covariance of the Gaussian distribution. These are randomly generated one time and fixed for the rest of the experiment.
    Simulated data are drawn from $p_{\text{miss}}$ while calibration data use the true prior. The experiment was carried out with  $\delta=2.0$ and $\alpha = 3.0$. Results are shown in Figure \ref{fig:prior_misspec_gaussian}. 
    \item\TaskC\ task : For this task, the true prior is set to an uniform distribution
    $ p(\rvtheta) = \mathcal{U}[0,45]$
    while the misspecified prior is set to a mixture between the true prior and a narrower  uniform distribution 
    $$p_{\text{miss}} = (1 - \tau) \, \mathcal{U}[0,45]) + \tau \, \mathcal{U}[10,20]$$
    where $\tau$ is  set to $\tau=1.0$. Other values for $\tau$ have been considered and yielded comparable results and where not included here.
    We used the same simulator (model A2C3) to generate the simulation and calibration data. Results are shown in Figure \ref{fig:prior_misspec_wt}.
\end{itemize}

\begin{figure}[h!]
    \centering
    \includegraphics[width=0.9\linewidth]{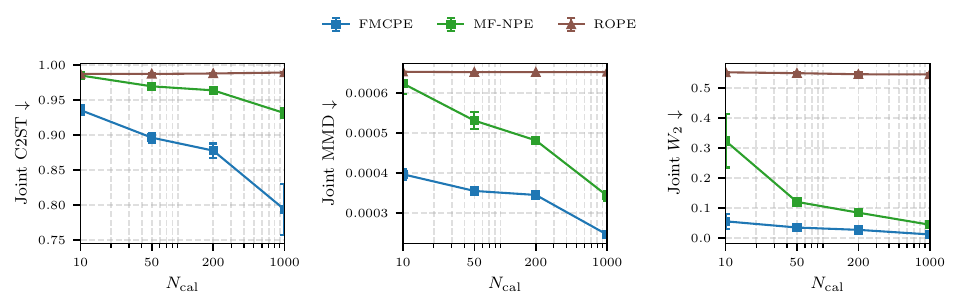}
    \caption{Method comparison under prior misspecification on the Gaussian task. Joint C2ST, MMD, and W2 between each method’s posterior samples and the reference posterior, with respect to the calibration set size . Means $\pm$ standard deviations are reported across 3 draws for the calibration set.}
    \label{fig:prior_misspec_gaussian}
\end{figure}
\begin{figure}
    \centering
    \includegraphics[width=0.9\linewidth]{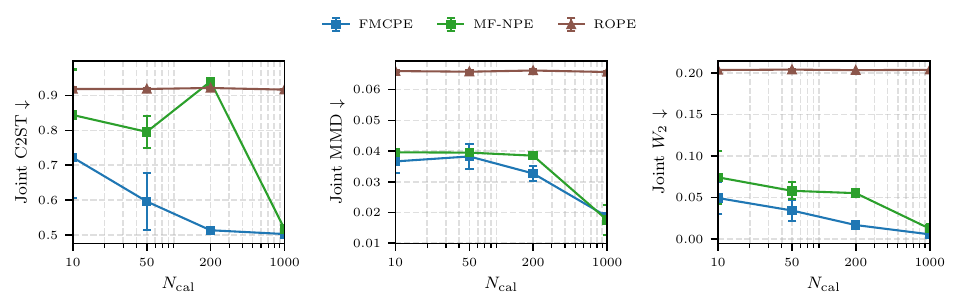}
    \caption{Method comparison under the most severe prior misspecification ($\tau = 1$) on the \TaskC~ task.
Joint C2ST, MMD, and W2 between each method’s posterior samples and the reference posterior, with respect to calibration size. Means $\pm$ standard deviations are reported across 3 draws of the calibration set}
    \label{fig:prior_misspec_wt}
\end{figure}

For both tasks,  our method gets better as $\ncal$ increases and outperforms the two other tested  baselines, \MFNPE\ and \ROPE . \ROPE\ does not seem to adapt to this particular type of misspecification, its performance does not increase with the size of the calibration set.
\section{Training and inference times}\label{app:timing}
We report the training (Table \ref{tab:training_rowstack}) and inference (Table \ref{tab:inference_rowstack}) times for our method and the baselines compared in Figure \ref{fig:boxplots}. Table \ref{tab:fm_component_breakdown_no_std} provides a detailed breakdown of the training time of our method across its main components. Although our method is more expensive to train than the baselines, it remains practical, requiring only 5–6 minutes per task. As shown in Table \ref{tab:fm_component_breakdown_no_std}, most of the training cost arises from the flow-sampling step (ODE integration), which is not optimized in our current implementation. We use a basic explicit-Euler solver, whereas more advanced integrators (e.g., midpoint methods, higher-order adaptive solvers, or JAX-based JIT compilation) could significantly reduce the number of integration steps and thus speed up training. Other methods such as distillation \citep{distillation_dao_2025} or straight flow trajectories \citep{fm_ot_one_step_2025} can help us reduce the training time even further. At inference time, our method is faster than \ROPE\ for all tasks and calibration sizes.

\begin{table}
\centering
\caption{Training time (in seconds) for each method and calibration set size $N_\text{cal}$. For the \TaskA\ task, no training time is reported for \ROPE\ as no  embedding network is used for NPE.}
\label{tab:training_rowstack}
\begin{adjustbox}{max width=\linewidth}
{\begin{tabular}{lcccc}
\toprule
Method & Gaussian & Light Tunnel & Pendulum & Wind Tunnel \\
\midrule

\multicolumn{5}{c}{\small \textbf{$N_\text{cal} = 10$}} \\
\midrule
NPE     & 5.51 & 0.79 & 0.63 & 0.58 \\
Ours    & 28.35 & 37.12 & 52.82 & 94.73 \\
MF-NPE  & 3.48 & 0.55 & 0.72 & 1.53 \\
RoPE    & --   & 5.40 & 7.81 & 7.10 \\
\\[-3pt]

\multicolumn{5}{c}{\small \textbf{$N_\text{cal}= 50$}} \\
\midrule
NPE     & 6.07 & 0.71 & 0.26 & 0.14 \\
Ours    & 63.29 & 70.97 & 80.74 & 126.74 \\
MF-NPE  & 6.33 & 1.76 & 0.76 & 1.92 \\
RoPE    & --   & 6.53 & 7.95 & 7.67 \\
\\[-3pt]

\multicolumn{5}{c}{\small \textbf{$N_\text{cal} = 200$}} \\
\midrule
NPE     & 13.11 & 4.11 & 2.21 & 0.19 \\
Ours    & 70.24 & 136.18 & 163.91 & 127.65 \\
MF-NPE  & 10.35 & 4.38 & 2.86 & 3.02 \\
RoPE    & --   & 15.93 & 11.87 & 11.73 \\
\\[-3pt]

\multicolumn{5}{c}{\small \textbf{$N_\text{cal}= 1000$}} \\
\midrule
NPE     & 37.32 & 15.10 & 12.32 & 5.94 \\
Ours    & 141.96 & 306.55 & 333.70 & 235.78 \\
MF-NPE  & 22.33 & 26.76 & 8.45 & 10.99 \\
RoPE    & --   & 61.57 & 38.09 & 38.47 \\
\bottomrule
\end{tabular}
}
\end{adjustbox}
\end{table}

\begin{table}
\centering
\caption{Inference time (in seconds)  for each method and calibration set size $N_\text{cal}$. For each method and task, we evaluated a test set of $N_{\text{test}} = 5000$ real-world observations $\{\bm{y}_j\}_{1\leq j \leq N_{\text{test}}}$. The reported times correspond to the total duration to generate all 5000 corresponding parameters.}
\label{tab:inference_rowstack}
\begin{adjustbox}{max width=\linewidth}
{
\begin{tabular}{lcccc}
\toprule
Method & \TaskA & \TaskD & \TaskB & \TaskC \\
\midrule

\multicolumn{5}{c}{\textbf{$N_\text{cal} = 10$}} \\
\midrule
NPE     & 0.09 & 0.67 & 0.12 & 0.02 \\
Ours    & 0.64 & 10.95 & 2.20 & 0.38 \\
MF-NPE  & 0.01 & 0.69 & 0.06 & 0.01 \\
RoPE    & 110.31 & 222.51 & 446.48 & 54.80 \\
\\[-3pt]

\multicolumn{5}{c}{\textbf{$N_\text{cal} = 50$}} \\
\midrule
NPE     & 0.01 & 0.63 & 0.07 & 0.01 \\
Ours    & 0.73 & 10.81 & 2.22 & 0.41 \\
MF-NPE  & 0.01 & 0.70 & 0.06 & 0.01 \\
RoPE    & 106.94 & 218.01 & 93.07 & 54.32 \\
\\[-3pt]

\multicolumn{5}{c}{\textbf{$N_\text{cal}= 200$}} \\
\midrule
NPE     & 0.01 & 0.64 & 0.07 & 0.01 \\
Ours    & 0.55 & 10.83 & 2.21 & 0.43 \\
MF-NPE  & 0.01 & 0.58 & 0.08 & 0.01 \\
RoPE    & 101.60 & 224.80 & 87.91 & 56.14 \\
\\[-3pt]

\multicolumn{5}{c}{\textbf{$N_\text{cal} = 1000$}} \\
\midrule
NPE     & 0.01 & 0.64 & 0.06 & 0.01 \\
Ours    & 0.40 & 11.03 & 2.20 & 0.43 \\
MF-NPE  & 0.01 & 0.65 & 0.08 & 0.01 \\
RoPE    & 100.11 & 226.29 & 88.23 & 59.13 \\
\bottomrule
\end{tabular}
}
\end{adjustbox}
\end{table}

\begin{table}
\centering
\caption{Training time (in seconds) for our method (\ourmethod) broken down into
3 components: (1) Gradient Step: time  to perform gradient updates on
the calibration set; (2) Sample NPE: time to sample from the base NPE
model; (3) Sample $\bm{x}$-flow: time  to sample from the learned flow in the
observation space.}
\label{tab:fm_component_breakdown_no_std}
\begin{adjustbox}{max width=\linewidth}
{
\begin{tabular}{llllll}
\toprule
 &  & $N_\text{cal}=10$ & $N_\text{cal}=50$ & $N_\text{cal}=200$ & $N_\text{cal}=1000$ \\
Task & Component &  &  &  &  \\
\midrule
\multirow[t]{3}{*}{\TaskA} 
 & Gradient Step      & 0.18 & 0.68 & 0.84 & 1.67 \\
 & Sample NPE         & 8.96 & 19.84 & 21.70 & 43.54 \\
 & Sample $\bm{x}$-flow    & 19.20 & 42.76 & 47.70 & 96.75 \\
\cline{1-6}
\multirow[t]{3}{*}{\TaskD} 
 & Gradient Step      & 0.68 & 1.43 & 2.36 & 2.81 \\
 & Sample NPE         & 9.82 & 18.24 & 33.68 & 58.25 \\
 & Sample $\bm{x}$-flow    & 26.62 & 51.29 & 100.14 & 245.49 \\
\cline{1-6}
\multirow[t]{3}{*}{\TaskB} 
 & Gradient Step      & 0.94 & 1.86 & 3.74 & 5.97 \\
 & Sample NPE         & 11.04 & 16.75 & 34.45 & 70.04 \\
 & Sample $\bm{x}$-flow    & 40.83 & 62.12 & 125.72 & 257.69 \\
\cline{1-6}
\multirow[t]{3}{*}{\TaskC} 
 & Gradient Step      & 1.10 & 1.68 & 1.84 & 2.90 \\
 & Sample NPE         & 21.03 & 27.93 & 27.91 & 52.04 \\
 & Sample $\bm{x}$-flow    & 72.60 & 97.13 & 97.90 & 180.83 \\
\cline{1-6}
\bottomrule
\end{tabular}
}
\end{adjustbox}
\end{table}

\section{Hyperparameter sensitivity}

%\subsection{Setup}

We conduct a grid search over the learning rate $\eta$, with $\eta \in \{10^{-4},\; 5\times10^{-4},\; 10^{-3},\; 5\times10^{-3}\}$, and over the gradient clipping threshold $g$, with $g \in \{0.5,\; 1.0,\; 5.0,\; \text{none}\}$, to assess the sensitivity of our method to these two commonly tuned hyperparameters. 
%All other settings are held fixed throughout.
%\paragraph{Grid search :}
%\begin{itemize}
%  \item Learning rate: $\eta \in \{10^{-4},\; 5\times10^{-4},\; 10^{-3},\; 5\times10^{-3}\}$
%  \item Gradient clipping: $g \in \{0.5,\; 1.0,\; 5.0,\; \text{none}\}$
%\end{itemize}
All other hyperparameters are fixed, we chose $N_{\text{epochs}}=50$ epochs and batch size $B=256$. We fixed $N_{\text{cal}}=200$, $N_{\text{sim}}=50{,}000$, $N_{\text{test}}=3{,}000$ and ran the grid search on 3 seeds $\in \{33,43,53\}$. We display a summary of the results in Table \ref{tab:hp_sensitivity} for the \TaskB\ task and in Table \ref{tab:hp_sensitivity_lt} for the \TaskD\ task. We also provide a visual representation of each metric sensitivity to the hyperparameters in Figures \ref{fig:hp_sensitivity_pendulum} and \ref{fig:hp_sensitivity_lt}.

%\paragraph{Metrics.} 
The metrics used are the average joint Wasserstein-2 ($jW_2$, $\downarrow$) and Joint MMD ($\downarrow$) distances $\pm$ std over 3 seeds.
On the \TaskB~task, performance varies meaningfully with the learning rate: $\eta = 10^{-3}$ yields the best results on both metrics, with $jW_2$ dropping by a factor of $2.5$ compared to the slowest rate $\eta = 10^{-4}$. The highest rate $\eta = 5\times10^{-3}$ leads to increased variance across seeds, suggesting mild instability. In contrast, gradient clipping has little effect across all learning rates. The four lines in Figure~\ref{fig:hp_sensitivity_pendulum} nearly overlap, indicating that the method is robust to this hyperparameter.

On the \TaskD\ task, $jW_2$ decreases more gradually and monotonically as the learning rate increases, with no clear optimum within the tested range. The jMMD metric is essentially flat across all 16 configurations, varying by less than $0.2\%$, which confirms that the method is highly stable on this task regardless of the hyperparameter choices made.

Overall, the learning rate has a moderate effect on $jW_2$ in both tasks, while gradient clipping appears largely inconsequential. We select $\eta = 10^{-3}$, $g = 1.0$ as our default configuration based on the Pendulum results.

% \subsection{Results: Pendulum}

\begin{table}[ht]
\centering
\caption{Hyperparameter  sensitivity (learning rate $\eta$ and gradient clipping $g$) on the \TaskB\ task ($N_{\text{cal}}=200$, mean $\pm$ std over seeds $\{33,43,53\}$). Best result per metric highlighted in bold \textcolor{best}{\textbf{green}}.}
\label{tab:hp_sensitivity}
\small
\begin{tabular}{cc cc}
\toprule
$\eta$ & $g$ &  $jW_2$ $\downarrow$ & jMMD $\downarrow$ \\
\midrule
\multirow{4}{*}{$10^{-4}$}
 & 0.5  & $6.43\times10^{-3}{}_{\pm 0.14\times10^{-3}}$ & $7.32\times10^{-4}{}_{\pm 0.07\times10^{-4}}$ \\
 & 1.0  & $6.45\times10^{-3}{}_{\pm 0.15\times10^{-3}}$ & $7.37\times10^{-4}{}_{\pm 0.07\times10^{-4}}$ \\
 & 5.0  & $6.38\times10^{-3}{}_{\pm 0.16\times10^{-3}}$ & $7.36\times10^{-4}{}_{\pm 0.11\times10^{-4}}$ \\
 & none & $6.39\times10^{-3}{}_{\pm 0.14\times10^{-3}}$ & $7.35\times10^{-4}{}_{\pm 0.06\times10^{-4}}$ \\
\midrule
\multirow{4}{*}{$5\times10^{-4}$}
 & 0.5  & $5.80\times10^{-3}{}_{\pm 0.39\times10^{-3}}$ & $7.28\times10^{-4}{}_{\pm 0.03\times10^{-4}}$ \\
 & 1.0  & $5.71\times10^{-3}{}_{\pm 0.41\times10^{-3}}$ & $7.27\times10^{-4}{}_{\pm 0.04\times10^{-4}}$ \\
 & 5.0  & $5.85\times10^{-3}{}_{\pm 0.37\times10^{-3}}$ & $7.28\times10^{-4}{}_{\pm 0.04\times10^{-4}}$ \\
 & none & $5.82\times10^{-3}{}_{\pm 0.37\times10^{-3}}$ & $7.28\times10^{-4}{}_{\pm 0.10\times10^{-4}}$ \\
\midrule
\multirow{4}{*}{$10^{-3}$}
 & 0.5  & $2.66\times10^{-3}{}_{\pm 0.41\times10^{-3}}$ & $6.97\times10^{-4}{}_{\pm 0.11\times10^{-4}}$ \\
 & 1.0  & \textcolor{best}{$\mathbf{2.60\times10^{-3}{}_{\pm 0.52\times10^{-3}}}$} & \textcolor{best}{$\mathbf{6.93\times10^{-4}{}_{\pm 0.16\times10^{-4}}}$} \\
 & 5.0  & $2.71\times10^{-3}{}_{\pm 0.33\times10^{-3}}$ & $6.95\times10^{-4}{}_{\pm 0.18\times10^{-4}}$ \\
 & none & $2.75\times10^{-3}{}_{\pm 0.30\times10^{-3}}$ & $7.02\times10^{-4}{}_{\pm 0.16\times10^{-4}}$ \\
\midrule
\multirow{4}{*}{$5\times10^{-3}$}
 & 0.5  & $4.53\times10^{-3}{}_{\pm 1.63\times10^{-3}}$ & $7.34\times10^{-4}{}_{\pm 0.09\times10^{-4}}$ \\
 & 1.0  & $6.44\times10^{-3}{}_{\pm 0.58\times10^{-3}}$ & $7.44\times10^{-4}{}_{\pm 0.13\times10^{-4}}$ \\
 & 5.0  & $4.76\times10^{-3}{}_{\pm 1.72\times10^{-3}}$ & $7.37\times10^{-4}{}_{\pm 0.09\times10^{-4}}$ \\
 & none & $3.85\times10^{-3}{}_{\pm 1.49\times10^{-3}}$ & $7.40\times10^{-4}{}_{\pm 0.08\times10^{-4}}$ \\
\bottomrule
\end{tabular}
\end{table}

% \subsection{Plots: Pendulum}

\begin{figure}[ht]
\centering
\begin{tikzpicture}
\begin{groupplot}[
  group style={group size=2 by 1, horizontal sep=2cm},
  width=0.47\textwidth, height=6cm,
  xlabel={Learning rate $\eta$},
  xmode=log,
  xtick={1e-4, 5e-4, 1e-3, 5e-3},
  xticklabels={$10^{-4}$, $5\!\times\!10^{-4}$, $10^{-3}$, $5\!\times\!10^{-3}$},
  legend style={font=\footnotesize, at={(0.98,0.98)}, anchor=north east},
  grid=major, grid style={gray!30},
  every axis plot/.append style={thick, mark size=2.5pt},
]

% Joint W2 vs LR, one line per grad clip
\nextgroupplot[ylabel={$jW_2$ $\downarrow$}, title={  $jW_2$ vs.\ learning rate}]
\addplot[blue,   mark=*]        coordinates {(1e-4,6.43e-3)(5e-4,5.80e-3)(1e-3,2.66e-3)(5e-3,4.53e-3)};
\addplot[red,    mark=triangle*] coordinates {(1e-4,6.45e-3)(5e-4,5.71e-3)(1e-3,2.60e-3)(5e-3,6.44e-3)};
\addplot[green!60!black, mark=diamond*] coordinates {(1e-4,6.38e-3)(5e-4,5.85e-3)(1e-3,2.71e-3)(5e-3,4.76e-3)};
\addplot[orange, mark=square*]  coordinates {(1e-4,6.39e-3)(5e-4,5.82e-3)(1e-3,2.75e-3)(5e-3,3.85e-3)};
\legend{$g=0.5$, $g=1.0$, $g=5.0$, $g=\text{none}$}

% Joint MMD vs LR, one line per grad clip
\nextgroupplot[ylabel={jMMD $\downarrow$}, title={jMMD vs.\ learning rate},
  scaled y ticks=false,
  yticklabel style={/pgf/number format/fixed, /pgf/number format/precision=4}]
\addplot[blue,   mark=*]        coordinates {(1e-4,7.32e-4)(5e-4,7.28e-4)(1e-3,6.97e-4)(5e-3,7.34e-4)};
\addplot[red,    mark=triangle*] coordinates {(1e-4,7.37e-4)(5e-4,7.27e-4)(1e-3,6.93e-4)(5e-3,7.44e-4)};
\addplot[green!60!black, mark=diamond*] coordinates {(1e-4,7.36e-4)(5e-4,7.28e-4)(1e-3,6.95e-4)(5e-3,7.37e-4)};
\addplot[orange, mark=square*]  coordinates {(1e-4,7.35e-4)(5e-4,7.28e-4)(1e-3,7.02e-4)(5e-3,7.40e-4)};

\end{groupplot}
\end{tikzpicture}
\caption{Hyperparameter sensitivity for the \TaskB\ task.  Joint $W_2$ and MMD vs.\ learning rate. Lines for different $g$ nearly overlap, demonstrating robustness with respect to gradient clipping.}
\label{fig:hp_sensitivity_pendulum}
\end{figure}

% \clearpage
% \subsection{Results: Light Tunnel}

\begin{table}[ht]
\centering
\caption{Hyperparameter sensitivity on the \TaskD\ task ($N_{\text{cal}}=200$, mean $\pm$ std over seeds $\{33,43,53\}$). Best result per metric in bold \textcolor{best}{\textbf{green}}.}
\label{tab:hp_sensitivity_lt}
\small
\begin{tabular}{cc cc}
\toprule
$\eta$ & $g$ &  $jW_2$ $\downarrow$ & jMMD $\downarrow$ \\
\midrule
\multirow{4}{*}{$10^{-4}$}
 & 0.5  & $5.29\times10^{-3}{}_{\pm 0.43\times10^{-3}}$ & $6.466\times10^{-4}{}_{\pm 0.004\times10^{-4}}$ \\
 & 1.0  & $5.26\times10^{-3}{}_{\pm 0.45\times10^{-3}}$ & $6.466\times10^{-4}{}_{\pm 0.006\times10^{-4}}$ \\
 & 5.0  & $5.21\times10^{-3}{}_{\pm 0.54\times10^{-3}}$ & $6.466\times10^{-4}{}_{\pm 0.006\times10^{-4}}$ \\
 & none & $5.30\times10^{-3}{}_{\pm 0.50\times10^{-3}}$ & $6.467\times10^{-4}{}_{\pm 0.007\times10^{-4}}$ \\
\midrule
\multirow{4}{*}{$5\times10^{-4}$}
 & 0.5  & $5.01\times10^{-3}{}_{\pm 0.65\times10^{-3}}$ & $6.474\times10^{-4}{}_{\pm 0.004\times10^{-4}}$ \\
 & 1.0  & $5.04\times10^{-3}{}_{\pm 0.60\times10^{-3}}$ & $6.476\times10^{-4}{}_{\pm 0.004\times10^{-4}}$ \\
 & 5.0  & $4.92\times10^{-3}{}_{\pm 0.56\times10^{-3}}$ & $6.474\times10^{-4}{}_{\pm 0.005\times10^{-4}}$ \\
 & none & $4.96\times10^{-3}{}_{\pm 0.67\times10^{-3}}$ & $6.475\times10^{-4}{}_{\pm 0.005\times10^{-4}}$ \\
\midrule
\multirow{4}{*}{$10^{-3}$}
 & 0.5  & $4.42\times10^{-3}{}_{\pm 0.15\times10^{-3}}$ & $6.472\times10^{-4}{}_{\pm 0.004\times10^{-4}}$ \\
 & 1.0  & $4.38\times10^{-3}{}_{\pm 0.11\times10^{-3}}$ & $6.472\times10^{-4}{}_{\pm 0.003\times10^{-4}}$ \\
 & 5.0  & $4.46\times10^{-3}{}_{\pm 0.12\times10^{-3}}$ & $6.472\times10^{-4}{}_{\pm 0.004\times10^{-4}}$ \\
 & none & $4.40\times10^{-3}{}_{\pm 0.20\times10^{-3}}$ & $6.471\times10^{-4}{}_{\pm 0.004\times10^{-4}}$ \\
\midrule
\multirow{4}{*}{$5\times10^{-3}$}
 & 0.5  & $4.06\times10^{-3}{}_{\pm 0.98\times10^{-3}}$ & $6.471\times10^{-4}{}_{\pm 0.009\times10^{-4}}$ \\
 & 1.0  & $4.03\times10^{-3}{}_{\pm 0.98\times10^{-3}}$ & $6.471\times10^{-4}{}_{\pm 0.008\times10^{-4}}$ \\
 & 5.0  & $4.51\times10^{-3}{}_{\pm 0.62\times10^{-3}}$ & $6.467\times10^{-4}{}_{\pm 0.012\times10^{-4}}$ \\
 & none & \textcolor{best}{$\mathbf{3.78\times10^{-3}{}_{\pm 1.18\times10^{-3}}}$} & \textcolor{best}{$\mathbf{6.468\times10^{-4}{}_{\pm 0.009\times10^{-4}}}$} \\
\bottomrule
\end{tabular}
\end{table}

% \subsection{Plots: Light Tunnel}

\begin{figure}[ht]
\centering
\begin{tikzpicture}
\begin{groupplot}[
  group style={group size=2 by 1, horizontal sep=3.2cm},
  width=0.47\textwidth, height=6cm,
  xlabel={Learning rate $\eta$},
  xmode=log,
  xtick={1e-4, 5e-4, 1e-3, 5e-3},
  xticklabels={$10^{-4}$, $5\!\times\!10^{-4}$, $10^{-3}$, $5\!\times\!10^{-3}$},
  legend style={font=\footnotesize, at={(0.98,0.98)}, anchor=north east},
  grid=major, grid style={gray!30},
  every axis plot/.append style={thick, mark size=2.5pt},
]

\nextgroupplot[ylabel={$jW_2$ $\downarrow$}, title={ $jW_2$ vs.\ learning rate}]
\addplot[blue,   mark=*]        coordinates {(1e-4,5.29e-3)(5e-4,5.01e-3)(1e-3,4.42e-3)(5e-3,4.06e-3)};
\addplot[red,    mark=triangle*] coordinates {(1e-4,5.26e-3)(5e-4,5.04e-3)(1e-3,4.38e-3)(5e-3,4.03e-3)};
\addplot[green!60!black, mark=diamond*] coordinates {(1e-4,5.21e-3)(5e-4,4.92e-3)(1e-3,4.46e-3)(5e-3,4.51e-3)};
\addplot[orange, mark=square*]  coordinates {(1e-4,5.30e-3)(5e-4,4.96e-3)(1e-3,4.40e-3)(5e-3,3.78e-3)};
\legend{$g=0.5$, $g=1.0$, $g=5.0$, $g=\text{none}$}

\nextgroupplot[ylabel={jMMD $\downarrow$}, title={jMMD vs.\ learning rate},
  scaled y ticks=false,
  ymin=6.46e-4, ymax=6.48e-4,
  yticklabel style={/pgf/number format/sci, /pgf/number format/precision=3}]
\addplot[blue,   mark=*]        coordinates {(1e-4,6.466e-4)(5e-4,6.474e-4)(1e-3,6.472e-4)(5e-3,6.471e-4)};
\addplot[red,    mark=triangle*] coordinates {(1e-4,6.466e-4)(5e-4,6.476e-4)(1e-3,6.472e-4)(5e-3,6.471e-4)};
\addplot[green!60!black, mark=diamond*] coordinates {(1e-4,6.466e-4)(5e-4,6.474e-4)(1e-3,6.472e-4)(5e-3,6.467e-4)};
\addplot[orange, mark=square*]  coordinates {(1e-4,6.467e-4)(5e-4,6.475e-4)(1e-3,6.471e-4)(5e-3,6.468e-4)};

\end{groupplot}
\end{tikzpicture}
\caption{Hyperparameter sensitivity for the \TaskD\ task: Joint $W_2$ and MMD vs.\ learning rate. $jW_2$ decreases monotonically with $\eta$ on this task. Joint MMD is essentially constant across all 16 configurations (range $< 0.2\%$), confirming extreme robustness.}
\label{fig:hp_sensitivity_lt}
\end{figure}
\end{document}